\documentclass[]{memtensor}

\usepackage[toc,page,header]{appendix}
\usepackage[utf8]{inputenc} % allow utf-8 input
\usepackage[T1]{fontenc}    % use 8-bit T1 fonts
\usepackage{hyperref}       % hyperlinks
\usepackage{url}            % simple URL typesetting
\usepackage{booktabs}       % professional-quality tables
\usepackage{amsfonts}       % blackboard math symbols
\usepackage{nicefrac}       % compact symbols for 1/2, etc.
\usepackage{microtype}      % microtypography
\usepackage{xcolor}         % colors
\usepackage{amsmath} 
\usepackage{etoolbox}
\usepackage{lipsum}  % 导入宏包
\usepackage{minitoc}
\usepackage[table]{xcolor}  % colors + table color support
\usepackage{tablefootnote}  % footnote in table
\usepackage{threeparttable}

\usepackage{algorithm}
\usepackage{algorithmic}
\usepackage{multicol}
\usepackage{fancyvrb}
\usepackage{listings}
\usepackage{xcolor}
\usepackage{tikz}
\usepackage{multirow}

\usetikzlibrary{positioning,fit,arrows.meta,calc}
\title{From Memory to Skills: Evidence-Grounded Co-Evolution Governance for Long-Horizon LLM Agents}

\author[1,2]{Bo Tang}
\author[3]{Yang Zhang}
\author[1]{Guomian Zhuang}
\author[1]{Wenqiang Wei}
\author[4]{Gaoyang Zheng}
\author[3]{Lindong Xie}
\author[4]{Yanchao Tan}
\author[1]{Feiyu Xiong}
\author[5]{Qingyu Yang}
\author[3]{Edward Chung}
\author[2]{Zhiyu li}

\affiliation[1]{MemTensor}
\affiliation[2]{University of Science and Technology of China}
\affiliation[3]{Hong Kong Polytechnic University}
\affiliation[4]{Fuzhou University}
\affiliation[5]{Xi'an Jiaotong University}

\abstract{
Existing memory systems for long-horizon LLM agents often retrieve prior traces as passive context rather than converting them into executable capabilities. In this paper, we propose MSCE, a training-free Memory--Skill Co-Evolution framework that organizes agent experience into grounded step traces, reusable procedural policies, and declarative environmental cognition. MSCE crystallizes evidence-backed L2 policies with positive estimated gain into callable skills that retain evidence links, applicability boundaries, decision guidance, verification rules, and reliability estimates. 
It further introduces reflection-weighted value backfilling, which propagates sparse terminal feedback through dense local self-reflections to produce evidence-calibrated trace values for governing memory and skill evolution.
Experiments on EvoAgentBench and LoCoMo demonstrate that MSCE significantly outperforms state-of-the-art skill-augmented and memory-driven agent baselines, exhibiting strong cross-domain transferability and lifelong-evolution capabilities. 
}
\date{\today}
\correspondence{Yang Zhang at \email{zynolo96@outlook.com}}
\checkdata[Code]{\url{https://github.com/MemTensor/MemOS}}

\begin{document}
\maketitle

\section{Introduction}

% Large Language Model (LLM) agents are increasingly deployed as interactive systems for long-horizon, multi-step, and user-specific tasks. Beyond generating responses, modern agents are expected to use tools, edit files, interact with external applications, and coordinate workflows across communication channels. 
% % Personal-agent platforms such as OpenClaw exemplify this shift, enabling agents to run on a user's machine, access local files and browsers, execute shell commands, and extend themselves through plugins~\citep{openclaw_personal_ai_assistant_2026}. 
% In such settings, long-term memory is essential for overcoming limited context windows: by storing, organizing, retrieving, and updating past interactions, agents can maintain coherence, personalize behavior, and reuse prior experience across sessions.

Large Language Model (LLM) agents are increasingly deployed for long-horizon, multi-step, and user-specific tasks involving tool use, file editing, application interaction, and workflow coordination. In such settings, long-term memory is essential for overcoming limited context windows, ensuring agents can maintain coherence, personalize behavior, and reuse experience across sessions.

% Existing memory mechanisms for LLM agents typically preserve either \textit{factual memory}, such as raw interactions, user facts, and compressed context~\citep{zhou2023recurrentgpt,xu2025mem,chen2025compress,chhikara2025mem0,kang2025memory}, or \textit{experiential memory}, such as reflections, reasoning traces, and task summaries~\citep{liu2023think,shinn2023reflexion,zhao2024expel,zhang2025memgen,ouyang2025reasoningbank}. Although useful for recall and guidance, these memories are usually reused as retrieved content rather than transformed into operational knowledge. As a result, long-horizon agents face two limitations. First, repeated traces are seldom distilled into executable action pipelines. For example, stored plugin-installation experiences that separately reveal manifest inspection, dependency resolution, registration, agent reload, and availability checks could form an ordered installation procedure, rather than be retrieved and reasoned over from scratch. Second, agents seldom build environmental cognition: even after repeatedly exploring a repository, they may still re-list directories to find components, tests, configuration files, or build scripts, leading to excessive token consumption.

Existing memory mechanisms typically store either \textit{factual memory}, such as raw interactions, user facts, and compressed contexts~\citep{zhou2023recurrentgpt,xu2025mem,chen2025compress,kang2025memory}, or \textit{experiential memory}, such as reflections, reasoning traces, and task summaries~\citep{liu2023think,shinn2023reflexion,zhao2024expel,zhang2025memgen}. These memories are often reused as passive context rather than transformed into operational knowledge. For example, repeated plugin-installation traces may reveal manifest inspection, dependency resolution, registration, reload, and availability checks, yet most systems still retrieve these traces and ask the agent to reason over them again. Moreover, after repeatedly exploring the same repository, an agent may still re-list directories to locate tests, configuration files, or build scripts, leading to excessive token consumption.

% Recently, skills provide an emerging solution. Unlike generic experience summaries that merely remind the agent of past observations or strategies, skills provide executable instructions that can be directly invoked, thus reducing repeated reasoning and improving the consistency of action generation~\citep{anthropic_agent_skills_2025}. However, turning agent trajectories into executable skills is non-trivial for three reasons. 
% First, raw interaction traces are noisy and often contain blind exploration, failed tool calls, or environment-specific artifacts. 
% Second, external feedback is usually sparse and delayed, making it difficult to decide which intermediate steps should be reinforced or avoided. 
% Third, reusable skills require not only a successful procedure, but also trigger conditions, applicability boundaries, and verification criteria. 
% % Therefore, directly distilled skills may overfit to individual trajectories and be invoked in inappropriate contexts.

Skills offer a more executable form of reusable knowledge: instead of merely reminding the agent of past observations, a skill provides an actionable procedure that guides future actions~\citep{anthropic_agent_skills_2025}. However, turning agent trajectories into reusable skills is difficult. Raw traces contain failed attempts, blind exploration, and environment-specific artifacts, while terminal feedback is often sparse and delayed, making step-level credit assignment uncertain. Moreover, a reusable skill requires not only a successful procedure, but also triggers, applicability boundaries, verification criteria, and lifecycle maintenance. Directly distilled skills may thus overfit to raw trajectories and be invoked in inappropriate contexts.

% To address these challenges, we propose MSCE, a \textbf{M}emory-\textbf{S}kill \textbf{C}o-\textbf{E}volution framework. The central idea is to govern memory as a hierarchy of increasingly abstract cognitive objects and to crystallize reliable memory into skills. Specifically, MSCE maintains three levels of memory: 
% \textit{L1 Trace Memory} records grounded, step-level execution loops, providing the foundational evidence to ensure that skills are derived from authentic interactions. \textit{L2 Policy Memory} induces reusable action strategies across different trajectories, serving as the structural skeleton (e.g., triggers, executable procedures, and verifications) for skills. \textit{L3 World Model} abstracts the spatial structures and behavioral rules of the environment, equipping the skills with environmental awareness and domain priors to guide exploration-free planning.
% %from accumulated L2 policies

To address these challenges, we propose MSCE, a training-free \textbf{M}emory-\textbf{S}kill \textbf{C}o-\textbf{E}volution framework that promotes skills from governed memory rather than directly distilling them from noisy trajectories. MSCE organizes experience into three memory levels. \textit{L1 trace memory} stores grounded step-level evidence. \textit{L2 policy memory} induces recurring procedural patterns from cross-episode traces. \textit{L3 environmental cognition memory} abstracts declarative knowledge about environmental structure and constraints. This hierarchy separates evidence, procedure, and environmental knowledge, converting noisy interaction histories into governable abstractions that can be verified, revised, and deployed as reusable skills.

% Building upon this governed memory, MSCE introduces an evidence- and value-guided skill crystallization mechanism, where a skill is crystallized only when an L2 policy demonstrates sustained positive utility and stability across multiple independent tasks. 
% % Unlike skills directly distilled from noisy raw trajectories~\citep{qiu2026autorefine,alzubi2026evoskill}, such memory-crystallized skills are filtered by value signals, grounded in evidence, and constrained by learned environmental priors, making them less brittle and more reliably reusable across recurrent tasks.
% To drive the continual evolution of both skills and the underlying memory, MSCE presents a two-fold feedback mechanism integrating \textit{step-level internal reflections} and \textit{task-level external feedback}. Since the former is dense but prone to hallucination, while latter is reliable but sparse, 
% MSCE employs reflection-weighted backpropagation to distribute task-level rewards across historical steps,
% estimating trace-level values. These values then serve as unified signals to drive parameter-free, autonomous refinement: they prioritize high-value traces during retrieval, weight the induction of new policies, and continuously update the reliability and decision boundaries of crystallized skills. 

On top of this governed memory, MSCE crystallizes useful and evidence-backed L2 policies into skills. 
A policy becomes eligible when it retains supporting evidence, exhibits positive estimated gain, and remains consistent with its current trigger, procedure, and applicability boundary.
The resulting skills preserve evidence anchors, verification rules, and a reliability-driven lifecycle.
To provide the value signal underlying this governance, MSCE introduces reflection-weighted value backfilling, which propagates sparse terminal feedback over grounded traces through local self-reflections. The resulting evidence-calibrated trace values serve as unified signals for retrieval, policy induction, environmental cognition abstraction, skill promotion, and later revision.

In summary, our main contributions are:

\begin{itemize}

\item \textbf{Framework:} We propose MSCE, a training-free memory--skill co-evolution framework that organizes long-horizon agent experience into grounded traces, procedural policies, and declarative environmental cognition.

\item \textbf{Method:} We design a governed memory-to-skill promotion mechanism that crystallizes only supported, positive-gain, and stable L2 policies into callable skills. We further introduce reflection-weighted value backfilling, which couples sparse terminal feedback with dense self-reflections to produce evidence-calibrated trace values for jointly updating memory and skills.

\item \textbf{Experiments:} Comprehensive evaluations on EvoAgentBench and LoCoMo show that MSCE achieves state-of-the-art performance. Further analyses validate its effectiveness in handling long-horizon and multi-domain environments, showcasing positive cross-domain transfer and lifelong-evolution gains.

\end{itemize}

\section{Related Work}

\subsection{Memory for LLM Agents}
%maintaining long-term coherence and personalization

Following~\citet{wang2026memex}, we broadly categorize existing long-term memory systems for LLM agents into two paradigms: factual memory and experiential memory.
\textit{Factual memory} focuses on retrieving raw interaction logs or observations to maintain consistency. Methods in this paradigm enhance retrieval mechanisms~\citep{chen2025compress,long2025seeing} or leverage knowledge organization, structuring memory into interconnected networks~\citep{xu2025mem}, neurobiologically inspired graphs~\citep{gutierrez2024hipporag,chhikara2025mem0}, or hierarchical storage~\citep{sun2026h}. While effective for recalling past events~\citep{zhou2023recurrentgpt,zhou2025memento}, they often overwhelm the context window with episodic noise and lack higher-level abstraction.

%Approaches like 
To address this, \textit{experiential memory} summarizes past trajectories to guide future reasoning. TiM~\citep{liu2023think} enables agents to learn from trial-and-error by storing evolving chains-of-thought or reflections. Architecture-driven frameworks have introduced OS-like systems~\citep{kang2025memory,hu2026evermemos} and advanced abstraction mechanisms~\citep{wang2026memex,zhangg,zhang2025memgen,ouyang2025reasoningbank} to dynamically manage context and distill reusable strategies. Despite these advances, most retrieved memories remain passive context and require the agent to reason over them again. MSCE differs by treating memory as a governed substrate from which procedural policies and deployable skills are induced.

% existing systems primarily serve as passive reminders of past contexts. They still require agents to perform heavy, repeated reasoning over retrieved summaries, falling short of crystallizing memories into structured, directly executable strategies. MSCE addresses this gap by leveraging memory not merely as a storage engine, but as a foundational substrate for continuous, autonomous skill crystallization.

\subsection{Skills for LLM Agents}

% Methods range from extracting workflows from successful trials~\citep{wangvoyager,wang2025agent} to employing continuous maintenance mechanisms~\citep{qiu2026autorefine,alzubi2026evoskill,yang2026autoskill,jiang2026xskill} or reinforcement learning~\citep{xia2026skillrl,wang2025reinforcement} for skill refinement.

Unlike generic experience summaries, skills are structured, self-contained packages encoding reusable procedures that enable agents to execute complex workflows directly. Recent skill acquisition methods generally follow two paradigms. \textit{External knowledge-based skills} rely on human expertise or curated datasets~\citep{chen2026cua,liang2026skillnet,jiao2026agentic}. For instance, CUA-Skill parameterizes human computer-use knowledge, while SkillNet constructs large-scale skill networks. However, acquiring high-quality expertise is costly, and these skills struggle to adapt to user-specific, open-ended environments. \textit{Interaction trajectory-based skills} enable autonomous adaptation by distilling the agent's own trial-and-error experiences~\citep{wangvoyager,wang2025agent,qiu2026autorefine,alzubi2026evoskill,yang2026autoskill,zhang2026skillflow}. For example, 
SkillFlow and EvoSkill introduce continuous maintenance mechanisms to extract and refine skills from execution trajectories. Besides, reinforcement learning approaches~\citep{xia2026skillrl,wang2025reinforcement} like SkillRL and SAGE optimize skills using environmental feedback.
Despite these efforts, directly distilling skills from ungoverned raw trajectories that are inherently noisy often yields brittle, over-specific procedures.
MSCE inserts an intermediate evidence-preserving memory hierarchy: L1 traces provide anchors, L2 policies aggregate cross-episode procedures, and L3 environmental cognition provides declarative context before skill deployment.

% By leveraging reflection-weighted value backfilling, MSCE biases later abstraction toward traces that are both positively rewarded and locally informative, while reducing the influence of unsupported or generic reflections.
%Early approaches like 
%three-level

%and XSkill~\citep{jiang2026xskill}

Concurrent works like MemSkill~\citep{zhang2026memskill} and ProcMEM~\citep{mi2026procmem} integrate memory and skills by recasting memory operations as skills or viewing skill pools as procedural memory. In contrast, MSCE separates three-level memory and executable skills, and links them through evidence anchors and value signals. This design turns memory--skill integration into a governed promotion problem: what becomes a skill, when it applies, and how it is revised or retired. We further discuss related work on self-evolving LLM agents in Appendix~\ref{app:sev_agents}.

% \begin{table*}[t]
% \centering
% \small
% \caption{Comparison between MSCE and representative memory/skill-based agent frameworks.}
% \label{tab:comparison}
% \begin{tabular}{lccccc}
% \toprule
% \textbf{Method} & \textbf{Trace Memory} & \textbf{Policy Abstraction} & \textbf{World Model} & \textbf{Skill Library} & \textbf{Feedback-based Revision} \\
% \midrule
% ReAct/Reflexion-like methods & \checkmark & partial & -- & -- & partial \\
% Experiential memory methods & \checkmark & \checkmark & partial & -- & partial \\
% Trajectory-to-skill methods & \checkmark & partial & -- & \checkmark & partial \\
% MemSkill / ProcMEM & \checkmark & \checkmark & --/partial & \checkmark & partial \\
% MSCE & \checkmark & \checkmark & \checkmark & \checkmark & \checkmark \\
% \bottomrule
% \end{tabular}
% \end{table*}

\section{Background}\label{sec:background}

In this paper, we formulate the LLM agent's interaction with a long-horizon environment across an ordered sequence of tasks. A task denotes a user-facing objective initiated by a query \(q\), while an episode denotes the feedback and update unit used by MSCE. In the common single-turn case, a task contains one episode; in multi-turn settings, a task may contain multiple episodes due to follow-up requests or corrections. We consider an ordered sequence of episodes \(\mathcal{E}=\{E_1,E_2,\ldots,E_n\}\). During episode \(E_i\), the agent generates \(H_i\) step traces $F_i = \{f_{i,1}, f_{i,2}, \ldots, f_{i,H_i}\}$.
At each step \(t\), the trace is represented as \((s_{i,t}, a_{i,t}, o_{i,t}, \rho_{i,t})\), where \(s_{i,t}\) denotes the semantic context, \(a_{i,t}\) is the primitive action, \(o_{i,t}\) is the observation returned by the environment, and \(\rho_{i,t}\) is the agent's self-reflection.

% In this paper, we formulate the LLM agent's interaction with a long-horizon environment across an ordered sequence of tasks \(\mathcal{T} = \{T_1, T_2, \ldots, T_n\}\). Each task \(T_i\) is initiated by a user query \(q_i\), which defines the task goal. 
% During the execution of task \(i\), the agent generates a sequence of \(H_i\) step traces 
% $f_i = \{f_{i,1}, f_{i,2}, \ldots, f_{i,H_i}\}$.
% At each step \(t\), the trace is represented as a tuple \((s_{i,t}, a_{i,t}, o_{i,t}, \rho_{i,t})\), where \(s_{i,t}\) denotes the current semantic context, \(a_{i,t}\) is the primitive action taken by the agent, \(o_{i,t}\) is the observation returned by the environment, and \(\rho_{i,t}\) is the agent's self-reflection over \(o_{i,t}\). 

Within this environment, the agent can leverage a crystallized skill library
\(\mathcal{K}\).
% , whose entries are structured skill objects containing an
% invocation guide, preconditions, ordered steps, examples, tool constraints,
% and evidence links.
% The library \(\mathcal{K}\) may be initially empty and is populated online by the MSCE update procedure introduced in Section~\ref{sec:method}.
Rather than exploring from scratch at every step, the agent monitors whether the current state \(s_{i,t}\) satisfies the trigger condition of any skill in \(\mathcal{K}\). If a relevant skill \(k = \mu(s_{i,t}, \mathcal{K})\) is retrieved, it is injected into the context to guide action generation: $a_{i,t} = \pi_{\text{LLM}}(\cdot \mid s_{i,t}, k)$.
Otherwise, the policy falls back to generating actions conditioned only on the current state. % the agent returns a task-level output \(\hat{y}_i\) to the user.
After episode \(E_i\) terminates, the system receives a terminal feedback signal. This signal may be a numerical reward from an environment verifier or textual feedback from the user. For textual feedback, we use the reward quantification mechanism in Appendix~\ref{app:reward_shaping} to convert it into a scalar terminal feedback \(R_i\). The primary objective is to maximize cumulative episode-level feedback \(\sum_{i=1}^{n} R_i\). In
practice, \(R_i\) may reflect task success, user satisfaction, process
quality, safety, or efficiency.

% therefore, token cost and redundant exploration can be reflected either through explicit verifier penalties or through the process-quality term in textual feedback quantification.

\section{Method}
\label{sec:method}

\subsection{Overview}

\begin{figure*}[!t]
\centering
\includegraphics[width=0.9\linewidth]{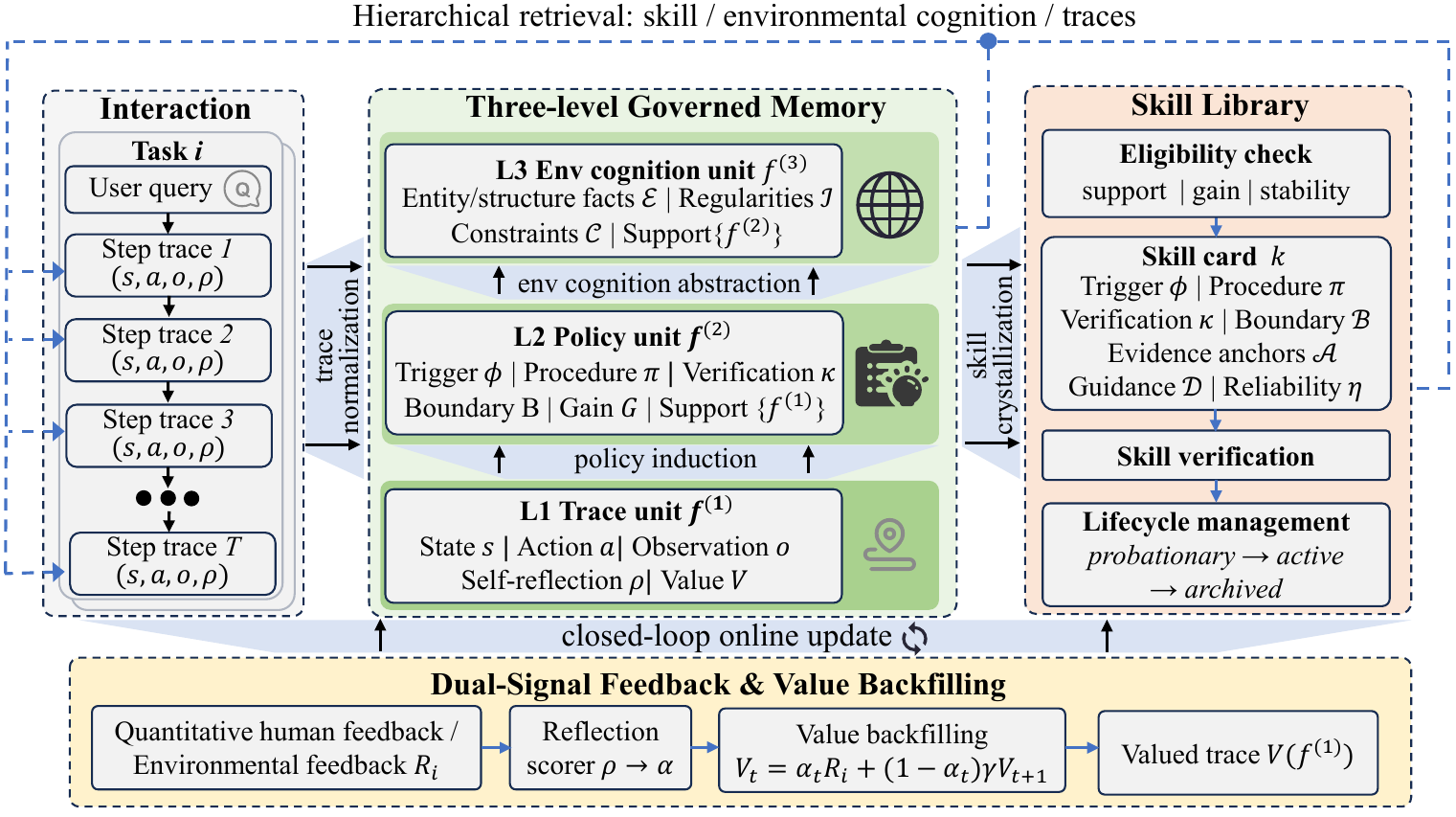}
\caption{Overview of MSCE with governed memory, skill crystallization, and dual-signal value backfilling.
}
\label{fig:msce_overview}
\vspace{-4mm}
\end{figure*}
% Overview of MSCE. An interaction task is normalized into grounded L1 step traces. Task-level feedback is combined with reflection scores to backfill step values, which govern a three-level memory hierarchy: valued L1 traces support L2 policy induction, and active policies support L3 world-model abstraction. Eligible L2 policies are crystallized into callable skills with evidence links, decision guidance, and reliability estimates. The resulting skills and memories are retrieved in later tasks, forming a closed-loop online update process.

% is a training-free framework that converts long-horizon interaction
% experience into reusable procedural knowledge and environment-level
% abstractions. It

The overview structure of MSCE is illustrated in Fig.~\ref{fig:msce_overview}. It maintains a memory hierarchy
\(\mathcal{M}=(\mathcal{M}^{(1)},\mathcal{M}^{(2)},\mathcal{M}^{(3)})\),
where L1 traces preserve auditable evidence, L2 policies abstract revisable
procedures, and L3 environmental cognition stores declarative environment knowledge.
The skill library \(\mathcal{K}\) is downstream of this memory hierarchy, exposing verified procedures for invocation. This separation keeps skills traceable to evidence while
preventing traces, policies, environmental facts, and callable actions from being
merged into a single uncontrolled pool. MSCE crystallizes evidence-backed
L2 policies with positive estimated gain into skills, and uses
reflection-weighted value backfilling to update memory and skills from
terminal feedback.

\subsection{Three-level Memory}

\paragraph{L1 Trace Memory.}
L1 stores grounded decision units collected during interaction:
\[
    f_{i,t}^{(1)} = (s_{i,t}, a_{i,t}, o_{i,t}, \rho_{i,t}, V(f_{i,t}^{(1)})),
\]
where \(s_{i,t}\), \(a_{i,t}\), \(o_{i,t}\), and \(\rho_{i,t}\) denote the semantic state, action, observation, and self-reflection at step \(t\). 
The trace value \(V(f_{i,t}^{(1)})\) is initialized as unavailable and filled after terminal feedback arrives, as described in Section~\ref{sec:feedback}.

L1 is the evidence layer of MSCE: higher-level memories and skills must maintain links to their supporting traces, so that induced procedures can be traced back to concrete interactions.

To bound storage and reduce privacy risk, MSCE does not persist unbounded raw observations. Instead, L1 stores normalized evidence: truncated state text, agent action text, structured tool-call metadata, and capped tool inputs and outputs. Higher-level memories store evidence identifiers rather than duplicating raw observations.

% \paragraph{Episodes and step segmentation.}
% We separate (i) an \emph{interaction episode}---the credit-assignment unit spanning one user-initiated execution cycle---from (ii) \emph{L1 steps} inside that episode. 
% When a new user message arrives, a relation classifier (lexical rules with LLM tie-breaking over the prior query and answer) labels the message as a \emph{revision} (reopen the current episode), \emph{follow-up} (close the prior episode and open a new one under the same session), or \emph{new task} (new session). 
% Within an episode, steps are segmented at user-turn boundaries after at least one assistant turn; consecutive user clarifications merge, and tool-only prefixes attach to the step's upstream state. 
% Episodes therefore align with user-visible request--execution cycles rather than individual tool calls. 
% At retrieval time, high-value steps from a matched past episode may be rolled up into an episode-level plan (Appendix~\ref{app:hierarchical_retrieval}).

\paragraph{L2 Policy Memory.}
L2 stores reusable policies induced from recurring trace evidence:
\[
    f^{(2)} = (\phi,\pi,\kappa,\mathcal{B},\{f^{(1)}\}),
\]
where \(\phi\) is the trigger condition, \(\pi\) is a natural-language
procedure, \(\kappa\) specifies verification or fallback criteria,
\(\mathcal{B}\) records applicability boundaries, and \(\{f^{(1)}\}\)
is the supporting L1 evidence set.

\textbf{Policy extraction.}
When a valued L1 trace \(f_{i,t}^{(1)}\) becomes available, MSCE first
checks whether its step value satisfies \(V(f_{i,t}^{(1)})\ge v_{\min}\) where $v_{\min}$ denotes the threshold for policy extraction.
Eligible traces are matched against existing L2 policies using a
combination of embedding similarity and structured evidence such as
domain tags, tool types, and normalized error signatures. A matched trace updates the corresponding policy. If no policy matches, the trace is placed into a candidate pool keyed by a deterministic pattern signature.
A new L2 policy is induced by the prompt \(\Pi_{\mathrm{policy}}\) only when the same signature bucket contains evidence from at least
\(n_{\min}\) distinct episodes.
%Thus, a single high-value trace can support an existing policy or enter the pool, but cannot by itself induce a new policy when \(n_{\min}>1\).

\textbf{Policy gain.}
For each touched policy, MSCE computes a heuristic utility gain over per-step values. Let \(S_{\mathrm{with}}\) be traces linked to the policy
and \(S_{\mathrm{without}}\) be other traces in the same evaluation pool.
The gain is 
\begin{equation}
\label{eq:policy_gain}
    G(f^{(2)}) =
    \bar{V}_{\mathrm{with}}
    -
    \bar{V}_{\mathrm{blend}}(S_{\mathrm{without}}).
\end{equation}
When \(|S_{\mathrm{with}}|\ge 3\), \(\bar{V}_{\mathrm{with}}\) is a
softmax-weighted mean over \(V(f)\); else it falls back to the
arithmetic mean. The without-side aggregate is an arithmetic mean shrunk toward a baseline. Appendix~\ref{app:l2_gain} gives the full
definition. We use \(G\) as a heuristic signal rather than a causal effect
estimate: it determines whether a policy remains active, becomes eligible
for skill crystallization, or is retired.

\paragraph{L3 Environmental Cognition Memory.}

L3 compresses environment-level cognition as
\[
    f^{(3)} = (\mathcal{E},\mathcal{I},\mathcal{C},\{f^{(2)}\}),
\]
where \(\mathcal{E}\), \(\mathcal{I}\), and \(\mathcal{C}\) are entity or structure facts, action--response regularities, and environmental constraints, each linked to supporting L2 policies $\{f^{(2)}\}$. 
Unlike L2 policies outlining how to solve recurring tasks, L3 describes how a particular environment or domain is organized. 
For example, repeated policies about editing frontend files, locating tests, and modifying configuration files may support a repository environmental cognition.

L3 abstraction is triggered by prompt \(\Pi_{\mathrm{env}}\) when multiple active policies share a domain or environment context. 
The abstraction step extracts declarative facts from the contributing policies and their evidence, while excluding imperative procedural instructions. 
% This separation prevents L3 from duplicating skills: L2 answers ``what to do'', whereas L3 answers ``how the environment is structured''. 
When a new L3 draft overlaps with an existing cognition, MSCE merges it; otherwise, it creates a new environmental cognition. 
When later evidence weakens support for a cognition, its confidence and
retrieval priority can be reduced; low-confidence cognitions are retained for
audit but excluded from default retrieval.

% rather than silently overwriting previous models.

% \begin{table}[t]
% \centering
% \small
% \caption{Three levels of memory in MSCE.}
% \label{tab:memory}
% \resizebox{\columnwidth}{!}{
% \begin{tabular}{llll}
% \toprule
% \textbf{Level} & \textbf{Object} & \textbf{Source} & \textbf{Role} \\
% \midrule
% L1 & grounded trace & step execution & evidence anchor \\
% L2 & sub-task policy & cross-episode induction & skill skeleton \\
% L3 & world model & abstraction over L2 & environmental prior \\
% \bottomrule
% \end{tabular}
% }
% \end{table}

\subsection{Skill Crystallization and Invocation}
\label{sec:skill}

A skill is a callable object derived from an eligible L2 policy:
\[
    k = (\phi,\pi,\kappa,\mathcal{B},\mathcal{A},\mathcal{D},\eta).
\]
It inherits the trigger, procedure, verification rule, and boundary from L2, adding evidence anchors \(\mathcal{A}\), decision guidance \(\mathcal{D}\), and reliability \(\eta\).  
% Here $\mathcal{A}=\{\mathcal{A}^+,\mathcal{A}^-\}$ separates supporting
% evidence $\mathcal{A}^+$ (positive anchors) from counter-evidence
% $\mathcal{A}^-$ (anti-patterns) used to bound applicability.
While L2 policies and skills share several fields, they serve different roles. An L2 policy is an internal abstraction for evidence aggregation and revision; it may be incomplete, unstable, or not directly exposed to the agent. A skill is a deployment object with a standardized invocation interface, lifecycle state, reliability estimate, evidence anchors, and decision guidance. 
Thus, crystallization is not merely formatting an L2 policy, but a
promotion step that makes a policy callable only after evidence, gain, and recent consistency checks.

% Skill promotion requires three conditions. 
% First, the source L2 policy must have sufficient support according to the
% L2 support statistic and its retained source.
% First, the source L2 policy must have sufficient cross-episode support from
% retained evidence traces.
% Second, its
% estimated gain \(G(f^{(2)})\) must exceed a threshold. Third, its trigger
% and boundary must be stable, meaning that recent evidence does not require
% substantial procedure rewrite or boundary shrinkage.
% When these conditions are met, MSCE constructs a support set from positive evidence and counter-evidence linked to the policy. 
% Positive evidence provides the common procedure; counter-evidence constrains the boundary and supplies anti-patterns.

Skill promotion is governed by two gates. 
First, the source L2 policy must retain supporting traces and satisfy the positive-gain condition \(G(f^{(2)})>\theta_G\) where $\theta_G$ is a threshold for skill promotion.
Second, it must be stable: recent evidence should fit the current trigger,
procedure, and boundary, rather than forcing a substantial rewrite of
\(\phi\), \(\pi\), or \(\mathcal{B}\). 
When these gates are satisfied, MSCE constructs a support set from positive
evidence and counter-evidence. Positive evidence induces the common
procedure, while counter-evidence constrains the boundary and yields
anti-patterns.

\paragraph{Crystallization and verification.}
The crystallization prompt \(\Pi_{\mathrm{skill}}\) conditions on the policy, evidence anchors \(\mathcal{A}\), a whitelist of tools observed in evidence, and any decision-guidance seeds already attached to the policy. 
It returns structured steps, parameters, and an initial decision guidance \(\mathcal{D}=(\text{preferences},\text{anti-patterns})\). 
A deterministic verifier checks each draft skill before insertion. It validates the required schema, including the skill name, preconditions, ordered steps, examples, tool list, and decision-guidance fields. It then enforces evidence grounding: cited evidence identifiers must come from the support set, and declared tools from the whitelist observed in evidence traces. Finally, two lightweight coverage tests ensure that the draft neither invents unsupported commands nor deviates from the retained trace evidence. Drafts that fail any check are discarded rather than exposed as callable skills.

\paragraph{Deployment and reliability.} 
After deployment, MSCE estimates skill reliability with a smoothed success rate. 
Let \(n_{\mathrm{pass}}\) and \(n_{\mathrm{trial}}\) denote successful and total invocations, respectively:
\[
    \eta =
    \frac{n_{\mathrm{pass}}+1}{n_{\mathrm{trial}}+2}.
\]
User correction or rejection decreases \(\eta\) and may shrink \(\mathcal{B}\), while repeated successes increase it. 
High-reliability skills enter active retrieval; low-reliability skills remain probationary or are archived. 
During inference, MSCE retrieves active skills first, but conditions their invocation on the memory hierarchy. If the context also matches an L3 cognition, retrieved L3 facts provide environmental priors that instantiate parameters and interpret applicability boundaries without overwriting the skill procedure. For example, a dependency-installation skill may choose package-manager commands according to whether the L3 cognition indicates Alpine, Debian, or macOS. If no skill applies or a skill fails, MSCE falls back to L1 traces and episode-level retrieval.

% \paragraph{Deployment and reliability.} Reliability is updated after deployment. 
% Let \(n_{\mathrm{pass}}\) and \(n_{\mathrm{trial}}\) denote the number of successful and total invocations. 
% MSCE estimates skill reliability with a smoothed success rate:
% \[
%     \eta =
%     \frac{n_{\mathrm{pass}}+1}{n_{\mathrm{trial}}+2}.
% \]
% Explicit user correction or rejection decreases \(\eta\) and may shrink \(\mathcal{B}\), while repeated successful invocations increase \(\eta\). 
% Skills with high reliability enter active retrieval; skills with low reliability remain probationary or are archived. 
% During inference, MSCE retrieves active skills first, but skill invocation is not isolated from the memory hierarchy. When the current context also matches an L3 world model, the retrieved L3 facts are injected as environmental priors alongside the skill. These priors do not overwrite the skill procedure; instead, they help instantiate parameters and interpret applicability boundaries. For example, a dependency-installation skill may choose different package-manager commands depending on whether the L3 model indicates an Alpine, Debian, or macOS environment. If no relevant skill applies or a skill fails, MSCE falls back to L1 traces and episode-level retrieval.

\subsection{Online Update via Dual-Signal Value Backfilling}
\label{sec:feedback}

MSCE does not update the base LLM. Instead, it updates
the external cognitive state
\[
    \mathcal{C}_i =
    (\mathcal{M}^{(1)},\mathcal{M}^{(2)},\mathcal{M}^{(3)},\mathcal{K})_i
\]
through two complementary feedback sources: dense step-level reflections
and sparse terminal feedback. Reflections are available at
most interaction steps but may be noisy or unfaithful; terminal feedback
is more reliable but delayed and sparse. MSCE couples these signals by
reflection-weighted value backfilling.

To distribute the sparse terminal feedback \(R_i\) to each step trace, the terminal step inherits this value: $V(f_{i,H_i}^{(1)})=R_i$.
For earlier steps, MSCE applies reflection-weighted value backfilling:
\begin{equation}\label{eq:r_i}
    V(f_{i,t}^{(1)}) =
    \alpha_{i,t}R_i +
    (1-\alpha_{i,t})\gamma V(f_{i,t+1}^{(1)}),
\end{equation}
where \(\gamma\in[0,1)\) is a discount factor and \(\alpha_{i,t}\in[0,1]\) is a reflection weight. 
Intuitively, high-\(\alpha\) steps provide locally informative evidence for the final outcome, while low-\(\alpha\) steps mainly inherit value from later steps. This biases memory construction toward steps that are both globally rewarded and locally interpretable. The weight \(\alpha\) is estimated via the prompt \(\Pi_{\mathrm{reflexion\_score}}\) based on the self-reflection \(\rho_{i,t}\) and its local context: faithful, concrete, causally informative, and transferable reflections receive higher weights, whereas empty, tautological, or unsupported ones receive zero or low weights.

% Intuitively, high-\(\alpha\) steps are treated as locally informative evidence for the final outcome, while low-\(\alpha\) steps mainly inherit value from later steps. This design biases memory construction toward steps that are both globally rewarded and locally interpretable. $\alpha$ is estimated via the prompt \(\Pi_{\mathrm{reflexion\_score}}\) based on the self-reflection \(\rho_{i,t}\) and its local context. 
% Reflections that are faithful, concrete, causally informative, and transferable receive higher weights; empty, tautological, or unsupported reflections receive zero or low weights. 
% % This design limits the impact of hallucinated reflections by requiring them to be grounded in the associated state, action, and observation. 

\paragraph{Online update.}
After an episode closes, \(\mathcal{U}\) executes asynchronously. It first persists normalized L1 traces and embeddings. Once terminal feedback \(R_i\) is available, MSCE backfills step values \(V\). The valued traces are then used to associate or induce L2 policies; active policies may trigger L3 abstraction, and eligible policies may crystallize into skills. Tool-failure bursts or explicit user corrections extend the decision-guidance field \(\mathcal{D}\), and the resulting repair context is injected on the next turn rather than interrupting the current action.

\paragraph{Additional details.}
Appendix~\ref{sec:example} provides a concrete example illustrating the MSCE workflow. Appendix~\ref{app:implementation_details} gives implementation details on agent reflection, skill lifecycle, memory retrieval, decision guidance, and storage and privacy controls. Appendix~\ref{app:prompts} lists the prompts used by MSCE.

% for L2 policy induction \(\Pi_{\mathrm{policy}}\), L3 world-model extraction \(\Pi_{\mathrm{world}}\), skill crystallization \(\Pi_{\mathrm{skill}}\), reflexion scoring \(\Pi_{\mathrm{ref}}\), and reward quantification \(\Pi_{\mathrm{reward}}\).

% Algorithm~\ref{alg:msce_update} summarizes the online update. 
% The update proceeds from evidence writing to value assignment, policy update, world-model abstraction, and skill maintenance. 
% Stages that require LLM generation are evidence-constrained and validated before being inserted into memory.

% \begin{algorithm}[t]
% \small
% \caption{MSCE online update}
% \label{alg:msce_update}
% \begin{algorithmic}[1]
% \REQUIRE episode \(E_i\), task output \(\hat{y}_i\), feedback \(R_i\), cognitive state \(\mathcal{C}_i\)
% \STATE extract grounded L1 steps \(F_i=\{f^{(1)}_{i,t}\}_{t=1}^{T_i}\)
% \STATE estimate reflection weights \(\alpha_{i,t}\) for steps with reflections
% \STATE backfill \(V(f^{(1)}_{i,t})\) using \(R_i\), \(\alpha_{i,t}\), and \(\gamma\)
% \STATE append valued traces to \(\mathcal{M}^{(1)}\)
% \STATE associate high-value traces with existing L2 policies
% \STATE induce or revise L2 policies from cross-episode evidence
% \STATE abstract or update L3 world models from active L2 policies
% \STATE promote, repair, or retire skills according to support, gain, and reliability
% \RETURN updated cognitive state \(\mathcal{C}_{i+1}\)
% \end{algorithmic}
% \end{algorithm}

\section{Experiments}
\label{sec:experiments}

% We organize the experiments around three questions:
% (1) whether MSCE improves self-evolution across heterogeneous agent domains,
% (2) whether the evolved memories and skills generalize across settings and over long horizons,
% and (3) which components are responsible for the gains.

% Due to space constraints, we focus the main text on tool-use and lifelong agent evaluations, and report additional long-dialogue memory, preference-alignment, and feedback-density studies in
% Appendix~\ref{app:additional_exp}.

\subsection{Experimental Settings}
\label{sec:exp_setup}

\paragraph{Benchmarks.}
Our primary evaluation uses EvoAgentBench,\footnote{https://evermind-ai.github.io/EvoAgentBench}
a multi-domain benchmark for long-horizon agent self-evolution.
It covers five agent domains: information retrieval based on BrowseComp-Plus~\cite{chen2025browsecomp},
mathematical reasoning based on OmniMath~\cite{gao2025omni},
software engineering based on SWE-Bench~\cite{jimenez2024swe},
code implementation based on LiveCodeBench~\cite{jain2025livecodebench},
and knowledge work based on GDPVal~\cite{patwardhan2025gdpval}.
We abbreviate them as IR, Math, SE, Code, and KW, respectively.
We additionally evaluate on LoCoMo~\cite{maharana2024evaluating} to test long-term dialogue memory;
we report it separately because it stresses memory consistency rather than tool-use skill execution.

\paragraph{Baselines.}
We compare MSCE with three categories of baselines.
First, \textit{memory-based agents}, including EverOS~\cite{hu2026evermemos}, Memento, and MemSkill, reuse past trajectories or stored cases but do not explicitly crystallize reusable skills.
Second, \textit{trajectory-to-skill agents}, including EvoSkill, OpenSpace~\cite{hkuds_openspace_2026}, and SkillFlow-Evolve, synthesize or refine skills from interaction trajectories.
SkillFlow-Evolve is a patch-based skill-evolution baseline adapted from SkillFlow~\cite{zhang2026skillflow}.
Finally, we include a \textit{vanilla agent} that uses the same base model and tools but does not persist long-term memory or construct skills.

\paragraph{Implementation.}
All methods are implemented on the OpenClaw v2026.5.7 runtime\footnote{https://openclaw.ai/} and receive identical tool access, interaction budgets, and task order.
We use GPT-5.2 as the agent backbone and GPT-4o for auxiliary prompted operators, including reflection scoring, L2 policy induction, and L3 environmental cognition abstraction.
For online methods, memory and skills are accumulated only from online evolution episodes without updating the base LLM, and are then reused on later evaluation. Hyperparameters are reported in Appendix (Table~\ref{tab:hparams}).

\paragraph{Metrics.}
We report \textbf{Pass@1} as the main effectiveness
metric for EvoAgentBench. Pass@1 is the macro-averaged per-task success rate: each trial receives a domain-specific reward~$r$, and is counted as successful if $r>\tau_{eval}$ (default $\tau_{eval}{=}0$; $\tau_{eval}{=}0.6$ for KW). We also report \textbf{cost}, which is measured by the mean number of interaction \textbf{turns} or character count (abbreviated as \textbf{chars}) per trial; see Appendix~\ref{app:settings} for full evaluation protocol. On LoCoMo, we report GPT-4o judge scores for four question types, together with overall judge score and F1.

\begin{table*}[t]
\centering
\small
\newcommand{\pass}[1]{\large #1}
\newcommand{\cost}[1]{\small #1}
\caption{
Main results on EvoAgentBench. We report Pass@1 (\%) and Cost across five domains. Cost is measured in \textbf{chars} for Math and \textbf{turns} for other domains. For Pass@1, the best results are shown in bold, and the second best results are underlined. Cost is reported as an efficiency reference and should be interpreted jointly with Pass@1, as lower cost may reflect premature termination or less extensive tool use.
}
\label{tab:main_results}
\resizebox{\textwidth}{!}{
\begin{tabular}{lcccccccccc}
\toprule
\multirow{2}{*}{\textbf{Method}} 
& \multicolumn{2}{c}{\textbf{Information Retrieval}} 
& \multicolumn{2}{c}{\textbf{Mathematical Reasoning}} 
& \multicolumn{2}{c}{\textbf{Software Engineering}} 
& \multicolumn{2}{c}{\textbf{Code Implementation}} 
& \multicolumn{2}{c}{\textbf{Knowledge Work}} \\
\cmidrule(lr){2-3}
\cmidrule(lr){4-5}
\cmidrule(lr){6-7}
\cmidrule(lr){8-9}
\cmidrule(lr){10-11}
& Pass@1 $\uparrow$ & Cost $\downarrow$
& Pass@1 $\uparrow$ & Cost $\downarrow$
& Pass@1 $\uparrow$ & Cost $\downarrow$
& Pass@1 $\uparrow$ & Cost $\downarrow$
& Pass@1 $\uparrow$ & Cost $\downarrow$ \\
\midrule
Vanilla Agent
& \pass{12.31} & \cost{6.9} 
& \pass{30.00} & \cost{1290.0} 
& \pass{\underline{38.46}} & \cost{41.2} 
& \pass{51.28} & \cost{3.0} 
& \pass{44.83} & \cost{13.3} \\

Memento 
& \pass{13.85} & \cost{9.5} 
& \pass{32.00} & \cost{1091.3} 
& \pass{\underline{38.46}} & \cost{37.7} 
& \pass{\underline{58.97}} & \cost{4.1} 
& \pass{46.55} & \cost{16.4} \\

EverOS 
& \pass{\underline{21.54}} & \cost{14.0} 
& \pass{\underline{43.00}} & \cost{1745.3} 
& \pass{26.92} & \cost{59.7} 
& \pass{48.72} & \cost{7.2} 
& \pass{\underline{48.28}} & \cost{19.1} \\

EvoSkill 
& \pass{12.31} & \cost{9.7} 
& \pass{29.00} & \cost{1185.9} 
& \pass{26.92} & \cost{37.9} 
& \pass{\textbf{{61.54}}} & \cost{3.9} 
& \pass{\underline{48.28}} & \cost{15.3} \\

OpenSpace 
& \pass{15.38} & \cost{8.1} 
& \pass{28.00} & \cost{1468.3} 
& \pass{30.77} & \cost{48.4} 
& \pass{56.41} & \cost{4.5} 
& \pass{31.03} & \cost{13.2} \\

MemSkill 
& \pass{15.38} & \cost{8.4} 
& \pass{29.00} & \cost{1103.4} 
& \pass{\underline{38.46}} & \cost{37.3} 
& \pass{56.41} & \cost{3.9} 
& \pass{46.55} & \cost{17.1} \\

SkillFlow-Evolve
& \pass{16.92} & \cost{8.7} 
& \pass{33.00} & \cost{1304.0} 
& \pass{34.62} & \cost{56.4} 
& \pass{56.41} & \cost{4.1} 
& \pass{46.55} & \cost{11.9} \\

\midrule
MSCE 
& \pass{\textbf{26.15}} & \cost{8.7} 
& \pass{\textbf{47.00}} & \cost{1140.9} 
& \pass{\textbf{53.85}} & \cost{40.8} 
& \pass{\textbf{61.54}} & \cost{2.0} 
& \pass{\textbf{53.45}} & \cost{15.0} \\

\bottomrule
\end{tabular}}
\vspace{-2mm}
\end{table*}

\subsection{Main Results}
\label{sec:exp_main}

\paragraph{Results on EvoAgentBench.}

As shown in Table~\ref{tab:main_results}, MSCE achieves the best or tied-best Pass@1 in all five domains.
Compared with the strongest non-MSCE baseline in each domain, it improves Pass@1 by 4.61 points on IR, 4.00 on Math, 15.39 on SE, and 5.17 on KW.
On Code, MSCE ties the best baseline, EvoSkill, at 61.54\% Pass@1 while reducing cost from 3.9 to 2.0 turns.
These results indicate that MSCE improves self-evolution across heterogeneous agent tasks.

The cost results show that these gains are not obtained by simply spending more inference budget.
When comparing against the strongest non-MSCE baseline by Pass@1, breaking ties by lower cost, MSCE reduces cost on IR, Math, Code, and KW, with especially large reductions on Math and Code.
The only exception is SE, where MSCE improves Pass@1 by 15.39 points but increases cost from 37.3 to 40.8 turns, consistent with the additional repository navigation, editing, and testing often required by successful software-engineering solutions.
Overall, MSCE achieves the strongest accuracy while maintaining favorable efficiency in most settings.

% The best results are shown in bold and the second-best results are underlined.
\begin{table}[t]
\centering
\small
\setlength{\tabcolsep}{3.0pt}
\renewcommand{\arraystretch}{1.08}
\caption{
Results on LoCoMo. We report LLM-judge (GPT-4o) scores for four question types, including single-hop, multi-hop, temporal reasoning, and open-domain questions, together with the overall judge score and overall F1 score. Higher is better for all metrics.
}
\label{tab:locomo_results}
% \resizebox{\columnwidth}{!}{
{
\begin{tabular}{lcccc|cc}
\toprule
\textbf{Method}
& \textbf{Single} 
& \textbf{Multi} 
& \textbf{Temp.} 
& \textbf{Open} 
& \textbf{Overall} 
& \textbf{F1} \\
\midrule
Vanilla Agent
& 28.18 & 21.63 & 12.15 & \textbf{39.58} & 24.35 & 25.95 \\

Memento
& 48.51 & 14.18 & 28.04 & 13.54 & 35.78 & 30.02 \\

EverOS
& 59.93 & 46.10 & 38.32 & \underline{29.17} & 50.97 & 41.58 \\

EvoSkill
& 42.45 & 27.30 & 19.94 & 16.67 & 33.38 & 25.51 \\

OpenSpace
& 54.93 & 37.59 & 34.89 & 21.88 & 45.52 & 36.94 \\

MemSkill
& 52.91 & 29.43 & 31.46 & 13.54 & 41.69 & 34.00 \\

SkillFlow-Evolve
& \underline{74.08} 
& \underline{46.81} 
& \underline{41.43} 
& 25.00 
& \underline{59.22} 
& \underline{48.71} \\

\midrule
MSCE
& \textbf{75.98} 
& \textbf{47.87} 
& \textbf{44.24} 
& 28.13
& \textbf{61.23} 
& \textbf{49.89} \\
\bottomrule
\end{tabular}}
\vspace{-2mm}
\end{table}

\paragraph{Long-dialogue memory results.}
Table~\ref{tab:locomo_results} reports LoCoMo results.
MSCE achieves the best overall judge score and F1, outperforming the strongest baseline, SkillFlow-Evolve, by 2.01 and 1.18 points, respectively.
It also obtains the highest scores on single-hop, multi-hop, and temporal reasoning questions, indicating that the governed L1--L2--L3 hierarchy helps preserve factual evidence while abstracting long-range dependencies.
EverOS performs best on open-domain questions excluding Vanilla Agent, but MSCE ranks second on this category and achieves the strongest aggregate performance.

\paragraph{Simulated human feedback.}
Appendix~\ref{app:simulated_human_feedback} further studies MSCE with LLM-simulated human feedback injected at the episode level. 
% The feedback is converted into scalar rewards by our reward quantification mechanism. 
The results show that the introduction of simulated human feedback improves Pass@1 on most EvoAgentBench domains, suggesting that richer textual feedback can provide useful learning signals for memory--skill evolution.

\subsection{Generalization Analysis}
\label{sec:generalization}

\begin{figure}[!t]
\centering
\includegraphics[width=0.6\linewidth]{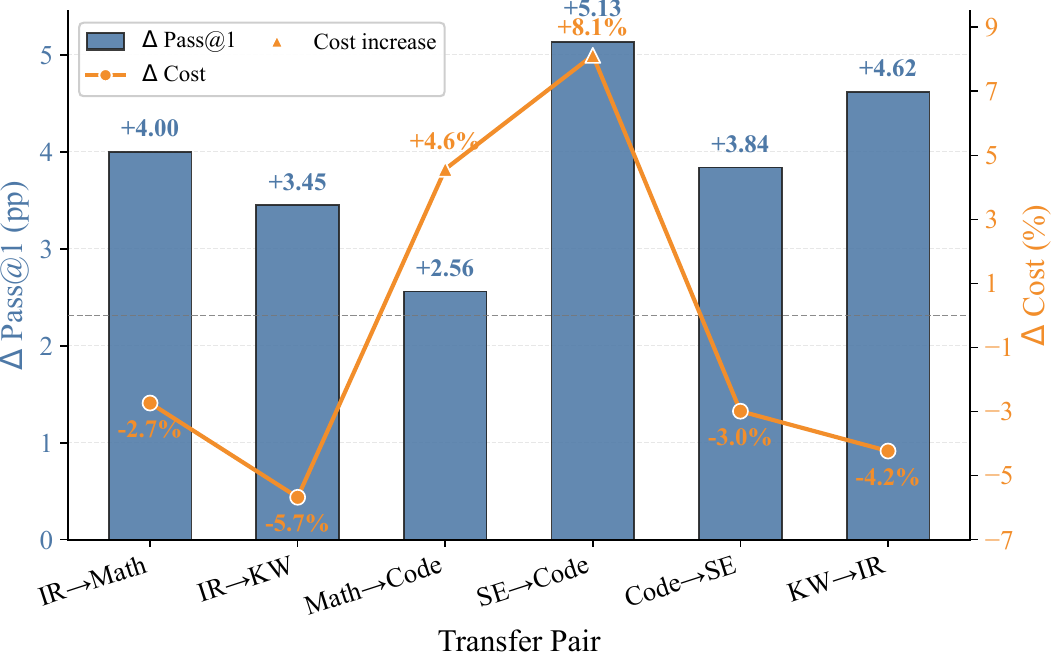}
\caption{
Cross-domain transfer results. 
% For each transfer pair \(A\rightarrow B\), \textit{intra} denotes evolution and evaluation on the target domain \(B\), while \textit{cross} reuses memory and skills evolved on the source domain \(A\) before evaluation on \(B\). 
Bars show the absolute Pass@1 improvement from \textit{intra} to \textit{cross} in percentage points, and the orange line shows the relative cost change.
}
\label{fig:cross_domain_transfer}
\vspace{-4mm}
\end{figure}

\begin{table*}[t]
\centering
\small
\newcommand{\pass}[1]{\large #1}
\newcommand{\cost}[1]{\small #1}
\caption{
Ablation results on EvoAgentBench. The top row shows the 
full MSCE method, while the following rows evaluate non-hierarchical and component-level ablations.
}
\label{tab:ablation}
\resizebox{\textwidth}{!}{
\begin{tabular}{lcccccccccc}
\toprule
\multirow{2}{*}{\textbf{Variant}} 
& \multicolumn{2}{c}{\textbf{Information Retrieval}} 
& \multicolumn{2}{c}{\textbf{Mathematical Reasoning}} 
& \multicolumn{2}{c}{\textbf{Software Engineering}} 
& \multicolumn{2}{c}{\textbf{Code Implementation}} 
& \multicolumn{2}{c}{\textbf{Knowledge Work}} \\
\cmidrule(lr){2-3}
\cmidrule(lr){4-5}
\cmidrule(lr){6-7}
\cmidrule(lr){8-9}
\cmidrule(lr){10-11}
& Pass@1 $\uparrow$ & Cost $\downarrow$
& Pass@1 $\uparrow$ & Cost $\downarrow$
& Pass@1 $\uparrow$ & Cost $\downarrow$
& Pass@1 $\uparrow$ & Cost $\downarrow$
& Pass@1 $\uparrow$ & Cost $\downarrow$ \\
\midrule
Full MSCE
& \pass{\textbf{26.15}} & \cost{8.7} 
& \pass{\textbf{47.00}} & \cost{1140.9} 
& \pass{\textbf{53.85}} & \cost{40.8} 
& \pass{\textbf{61.54}} & \cost{2.0} 
& \pass{\textbf{53.45}} & \cost{15.0} \\

\midrule
Flat Memory
& \pass{10.77} & \cost{9.0}
& \pass{31.00} & \cost{1322.2}
& \pass{34.62} & \cost{56.0}
& \pass{56.41} & \cost{5.3}
& \pass{37.93} & \cost{17.7} \\

w/o L3 
& \pass{23.08} & \cost{11.1}
& \pass{43.00} & \cost{1461.3}
& \pass{50.00} & \cost{54.6}
& \pass{58.97} & \cost{2.8}
& \pass{48.28} & \cost{19.5} \\

w/o Value Calibration
& \pass{24.62} & \cost{9.7}
& \pass{44.00} & \cost{1297.3}
& \pass{50.00} & \cost{46.4}
& \pass{58.97} & \cost{2.3}
& \pass{50.00} & \cost{16.4} \\

w/o Reflection Weighting
& \pass{21.54} & \cost{10.8}
& \pass{40.00} & \cost{1378.2}
& \pass{46.15} & \cost{50.7}
& \pass{56.41} & \cost{2.5}
& \pass{46.55} & \cost{17.6} \\

w/o Skill Crystallization
& \pass{20.00} & \cost{12.6}
& \pass{39.00} & \cost{1568.6}
& \pass{42.31} & \cost{58.8}
& \pass{53.85} & \cost{3.1}
& \pass{44.83} & \cost{20.3} \\
\bottomrule
\end{tabular}}
\vspace{-2mm}
\end{table*}

\paragraph{Cross-domain transfer.}
We evaluate whether evolved memory and skills transfer across domains: for each pair \(A\rightarrow B\), \textit{intra} evolves and evaluates the agent on target domain \(B\), whereas \textit{cross} initializes it with memory and skills evolved on source domain \(A\) before further evolution and evaluation on \(B\).

Figure~\ref{fig:cross_domain_transfer} shows that cross-domain transfer improves Pass@1 across all six transfer pairs, with gains from \(+2.56\) to \(+5.13\) points and an average gain of \(+3.93\) points, with cost decreasing in four of six pairs.
The two transfers into Code increase cost by \(+4.6\%\) and \(+8.1\%\), but also yield clear Pass@1 gains, suggesting a favorable accuracy--cost trade-off. These results indicate that MSCE learns transferable problem-solving structure rather than domain-specific artifacts.

\paragraph{Lifelong Evolution}

\begin{figure}[!t]
\centering
\includegraphics[width=0.7\linewidth]{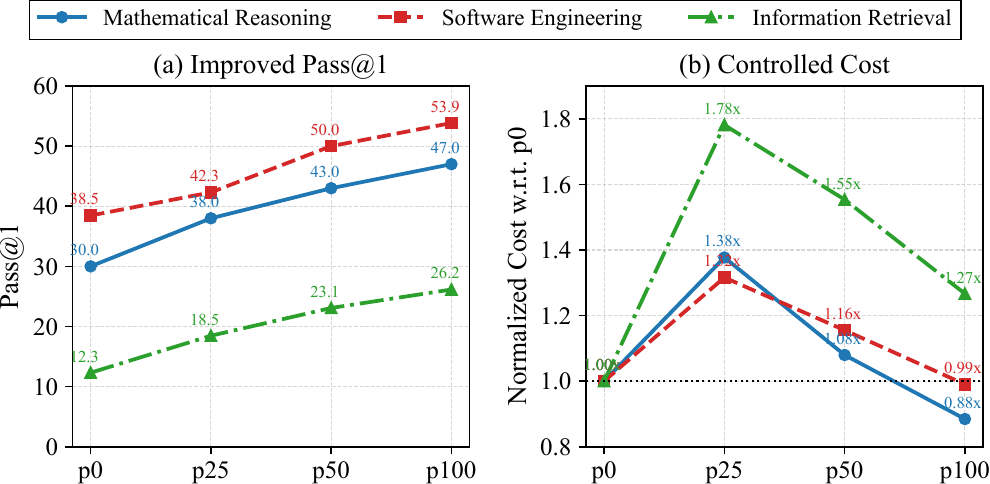}
\caption{
Long-horizon cumulative evolution under the lifelong learning protocol. The left panel shows Pass@1 across increasing accumulated training scales, and the right panel shows cost normalized by the corresponding p0 cost of each task category.} 
% Across all categories, Pass@1 improves monotonically from p0 to p100, while normalized cost first increases and then decreases, indicating that accumulated experience improves task success without simply increasing inference cost.}
\label{fig:long_horizon_evolution}
\vspace{-4mm}
\end{figure}

Figure~\ref{fig:long_horizon_evolution} presents long-horizon evolution under the lifelong learning protocol, where \(p0\) denotes the initial no-accumulation setting and \(p25\), \(p50\), and \(p100\) denote increasing accumulated experience scales.
Pass@1 improves monotonically from \(p0\) to \(p100\), with gains of 17.00, 15.39, and 13.84 points on Math, SE, and IR, respectively.
Normalized cost first increases at \(p25\) and then consistently decreases toward \(p100\).
At \(p100\), cost becomes lower than \(p0\) on Math and SE and is substantially reduced from its \(p25\) peak on IR.
These results support a ``learning by using'' effect: accumulated experience improves success without simply increasing inference cost.

\subsection{Ablation Study}

We ablate five components of MSCE.
\textbf{Flat Memory} stores all experiences in a single non-hierarchical memory and retrieves them directly at test time, without hierarchical abstraction, applicability gating, or skill introduction.
\textbf{w/o L3} removes L3 environmental cognition abstraction.
\textbf{w/o Value Calibration} directly injects retrieved skills without value-calibrated applicability filtering.
\textbf{w/o Reflection Weighting} removes reflection-weighted value backfilling and treats feedback uniformly.
\textbf{w/o Skill Crystallization} prevents distilled task policies from being converted into reusable skills.

Table~\ref{tab:ablation} shows that all components contribute to performance and efficiency.
The largest degradation comes from replacing MSCE with \textbf{Flat Memory}: Pass@1 drops by 15.38, 16.00, and 19.23 points on IR, Math, and SE, respectively, and Code cost increases from 2.0 to 5.3 turns despite a relatively small Pass@1 drop.
This indicates that merely retrieving past experiences is insufficient; effective reuse requires hierarchical abstraction, applicability control, and skill-level consolidation.

Among component-level ablations, removing skill crystallization causes the most consistent degradation, reducing Pass@1 by 6.15--11.54 points across domains and increasing cost for every domain.
Removing the L3 environmental cognition, reflection weighting, or the value calibration also consistently hurts performance.
In particular, disabling the value calibration lowers Pass@1 and increases cost in every domain, supporting the need for applicability-aware filtering to avoid harmful skill injection.
Overall, the ablations show that MSCE's gains arise from the joint effect of hierarchical organization, skill crystallization, calibrated selection, and reflection-weighted learning.

\section{Conclusion}

% In this paper, we introduced MSCE, a training-free memory--skill co-evolution framework for long-horizon LLM agents. By organizing experience into grounded traces, reusable policies, and declarative world models, and crystallizing evidence-backed positive-gain policies into callable skills, MSCE achieves state-of-the-art performance on both self-evolution and memory for LLM agents benchmarks, with significant cross-domain transfer and lifelong-evolution performance. We hope MSCE provide a solid foundation for connecting memory and skills for long-horizon LLM agents, contributing more efficient agent self-evolving.

In this paper, we introduced MSCE, a training-free memory--skill co-evolution framework for long-horizon LLM agents. By organizing experience into grounded traces, reusable policies, and environmental cognition, and by crystallizing evidence-backed, positive-gain policies into skills, MSCE bridges the gap between passive memory retrieval and active skill execution. Extensive experiments validate that MSCE achieves state-of-the-art performance on long-horizon self-evolution and conversational memory benchmarks, alongside significant cross-domain transfer and lifelong-evolution capabilities. We hope MSCE provides a foundation for connecting memory and skills, contributing to the development of more efficient and autonomously evolving LLM agents.

\section*{Limitations}

MSCE has several limitations. First, its value estimates, policy gains, and reliability scores are heuristic governance signals rather than causal credit-assignment estimates. They support ranking, filtering, and revision of memory objects, but do not guarantee that a promoted policy would cause success under counterfactual intervention. 

Second, several update operators, including reflection scoring, policy induction, environmental cognition abstraction, reward quantification, and skill drafting, rely on prompted LLMs. 
Although outputs are evidence-constrained and verified before insertion, they may still be noisy or sensitive to model and prompt choices, and they introduce additional latency and cost.

Third, our implementation depends on a particular agent runtime, tool interface, and backbone configuration; absolute results may vary under different models or environments.

Finally, MSCE stores normalized traces, evidence identifiers, policies, environmental cognition, and skills.  Although we apply truncation, deduplication, and rule-based redaction to reduce storage and privacy risks, these controls do not eliminate the possibility of retaining sensitive user information, environment-specific secrets, or unsafe procedural knowledge. Practical deployments should include stronger privacy filters, access control, and safety checks for generated skills.

% \section*{Acknowledgments}


\clearpage

\bibliographystyle{plainnat}
\bibliography{main}

\clearpage

\beginappendix
% \section{Appendix}

\section{More Related Work} 

\subsection{Self-Evolution of LLM Agents}\label{app:sev_agents}

Self-evolving LLM agents improve future behavior from past interactions through dual-signal feedback: \textit{internal reflection} and \textit{external task feedback}. Reflection-based methods, such as Self-Refine~\citep{madaan2023self}, Reflexion~\citep{shinn2023reflexion}, ExpeL~\citep{zhao2024expel}, SAMULE~\citep{ge2025samule}, and PreFlect~\citep{wang2026preflect}, generate natural-language critiques over outputs, trajectories, or plans to guide future attempts. While dense and interpretable, such reflections can be generic, mislocalized, or hallucinated. In contrast, feedback-driven methods such as LATS~\citep{zhou2024language}, LEAFE~\citep{ge2026internalizing}, and RetroAgent~\citep{zhang2026retroagent} exploit environment signals, verifiers, or rewards to guide search, distillation, or policy learning. These signals are more reliable but often sparse and delayed, making credit assignment difficult.

MSCE explicitly couples these two feedback sources through reflection-weighted value backfilling.
This produces unified trace-level values that guide memory retrieval, policy induction, and skill crystallization, enabling memory and skills to co-evolve in a training-free manner. 

% MSCE is not intended to introduce each primitive operation in isolation.
% Rather, its contribution lies in organizing reflection scoring, value
% backfilling, policy induction, world-model abstraction, and skill
% maintenance into a unified online governance pipeline, where all higher-level
% objects remain grounded in trace-level evidence and are revised through
% shared value signals.

\section{Implementation Details of MSCE}
\label{app:implementation_details}

This appendix provides implementation details omitted from the main text. 

\subsection{Agent Reflection}\label{app:reflection_generation}

The reflection field \(\rho_{i,t}\) is obtained from the agent runtime when
native reflective metadata is available; otherwise MSCE extracts inline reflective statements from the assistant turn, and may synthesize a short reflection with an LLM when no reflection is present. The reliability of \(\rho_{i,t}\) is not assumed, but later estimated through the reflection scoring procedure in Appendix~\ref{app:reflection_score}.

\subsection{Episode Definition and Boundary Management}
\label{app:episode_management}

The main text defines an \textit{episode} as the feedback and update unit
used by MSCE. This appendix provides the boundary-management details needed
to instantiate that definition in online multi-turn interaction. An episode
is not an additional memory level; rather, it determines when L1 traces are
closed, when terminal feedback is backfilled, and when downstream memory
updates are triggered.

\paragraph{Task, episode, and step.}
A \textit{task} is the user-facing objective introduced in Section~\ref{sec:background}, an
\textit{episode} is the feedback and update unit, and a \textit{step} is the
minimal L1 memory unit \(f_{i,t}^{(1)}\). Each episode closes a set of L1
steps that are assigned terminal feedback together. In the common
single-turn case, a task and an episode coincide. In multi-turn settings, a
task may contain multiple episodes when the user provides follow-up requests
or corrections. MSCE uses episode boundaries to decide whether a new
interaction should revise the previous evidence, continue the same
session, or start a new task context.

Formally, we write an episode as
\[
    E_i = \{\tau_{i,1}, \tau_{i,2}, \ldots, \tau_{i,H_i}\},
\]
where each \(\tau_{i,t}\) is an assistant response segment or a
tool-mediated action--observation segment. The capture module converts
\(E_j\) into an ordered set of L1 traces
\[
    F_i^{(1)} =
    \{f_{i,1}^{(1)}, f_{i,2}^{(1)}, \ldots, f_{i,H_i}^{(1)}\}.
\]
All traces in \(F_i^{(1)}\) share the same episode identifier, which is used
for reward backfilling, L2 evidence grouping, and episode-level retrieval.

\paragraph{Episode boundaries.}
MSCE maintains sessions across multiple user turns. When a new user message
arrives, a lightweight relation classifier determines how it relates to the
previous episode. The classifier combines lexical cues with an LLM-based
tie-breaker when necessary and returns one of three operational labels:
\begin{itemize}
    \item \textbf{Correction}: the new message corrects or rejects the previous
    result. MSCE extends the last episode so that the new evidence can
    revise the same credit-assignment unit.
    \item \textbf{Follow-up}: the new message extends the same goal or
    environment without invalidating the previous result. MSCE closes the
    previous episode and opens a new episode under the same task context.
    \item \textbf{New task}: the new message introduces an unrelated goal or
    domain. MSCE starts a new task context while keeping long-term
    memory retrievable.
\end{itemize}
This design separates the user-facing notion of a task from the system-level
unit used for online updates. It also avoids treating every tool call as an
independent task, which would make credit assignment too fine-grained and
unstable.

\paragraph{Role in value backfilling.}
Episode boundaries define the scope of credit assignment. Once feedback for
an episode is available, the traces in \(F_i^{(1)}\) are valued together
using the reflection-weighted backfilling rule in
Eq.~\ref{eq:r_i}. The same terminal feedback signal is shared
within the episode, while each step receives a different value according to
its reflection weight and temporal position. This is why MSCE does not treat
each tool call as an independent training instance: doing so would make the
feedback signal too sparse and unstable.

\paragraph{Role in L2 induction.}
Episodes also provide the independence criterion for policy induction. A
single high-value trace may be associated with an existing L2 policy or
placed into the candidate pool, but a new L2 policy is induced only when a
candidate signature accumulates evidence from at least \(n_{\min}\) distinct
episodes. This prevents multiple traces from the same execution trajectory
from being counted as independent support. In other words, the threshold for
L2 induction is cross-episode support, whereas the value threshold
\(V(f)\ge v_{\min}\) is applied to individual L1 traces.

\paragraph{Role in retrieval.}
Although retrieval can return individual L1 traces, MSCE may also aggregate
retrieved traces by episode. This episode-level rollup provides a compact
reference trajectory when the current context resembles a previous execution
segment, while still preserving the underlying trace-level evidence links.

% \paragraph{Why episode is not introduced in the main text.}
% We reserve the term \textit{episode} for implementation details because the
% main algorithm can be understood using only two conceptual units: the
% user-facing task and the step-level trace. Introducing episode in the main
% text would add a third granularity that is useful for implementation but not
% necessary for understanding the memory hierarchy itself. The appendix
% therefore makes the operational mapping explicit: tasks describe user
% objectives, steps define L1 evidence, and episodes define the online update
% boundary between them.

% \textcolor{red}{Better to show the text prompt of LLM evaluator}

% Compared to binary success/failure metrics or simple keyword matching, this multi-dimensional scoring mechanism provides a fine-grained terminal signal. It effectively captures nuanced user intents, allowing the system to distinguish between a ``completely incorrect'' execution (\(R_i\) close to \(-1\)) and a ``directionally correct execution that requires minor adjustments'' (\(R_i\) slightly greater than \(0\)). This quantified \(R_i\) is then utilized by the main framework to perform credit assignment across the trajectory \(f_i\).

\subsection{Reward Quantification from Textual Feedback}
\label{app:reward_shaping}

When no numerical environment reward is available, MSCE converts textual
user feedback into a normalized scalar \(R_i\in[-1,1]\). Given user
feedback \(h_i\) and a compact task summary, an LLM evaluator scores the
execution trajectory along three axes:
\begin{itemize}
    \item \textbf{Goal achievement} \(g_i\in[-1,1]\): whether the agent
    addressed the user's actual request, with later turns weighted more
    strongly in multi-turn tasks.
    \item \textbf{Process quality} \(p_i\in[-1,1]\): whether the execution
    was efficient, safe, and avoided unnecessary tool use or exploration.
    \item \textbf{User satisfaction} \(u_i\in[-1,1]\): whether the user's
    feedback expresses acceptance, neutrality, correction, or rejection.
\end{itemize}
The terminal reward is computed as
\[
    R_i =
    \mathrm{clip}\left(
    0.45 g_i + 0.30 p_i + 0.25 u_i,\,-1,\,1
    \right).
\]
If the LLM evaluator is unavailable, MSCE falls back to a heuristic
polarity-based estimate from the explicit feedback channel.

\subsection{Policy Gain for L2}
\label{app:l2_gain}

\paragraph{Sets.}
L2 policy gain is recomputed whenever reward backfilling finishes for an episode. For each touched policy, we form a finite trace pool
\(\mathcal{T}\) consisting of traces from the triggering episode and any
additional traces that served as cross-episode induction evidence for that
policy. The with set is $S_{\mathrm{with}} = \{\, f \in \mathcal{T} \mid f \text{ is linked to the policy in this update} \,\}$
including traces matched by association and traces in the inducing bucket.
The without set is $S_{\mathrm{without}} = \mathcal{T}\setminus S_{\mathrm{with}}$.
Earlier in the pipeline, only traces with \(V(f)\ge v_{\min}\) enter
association or the candidate pool; gain computation itself uses the stored
per-step values \(V(f)\).

\paragraph{With-side aggregate.}
For \(|S_{\mathrm{with}}|\ge 3\), MSCE uses a value-weighted mean:
\begin{equation}
\label{eq:l2_mu_w}
\mu_{\mathrm{w}}(S_{\mathrm{with}})
=
\sum_{f\in S_{\mathrm{with}}} w_f V(f),
\end{equation}
where
\begin{equation}
\label{eq:l2_softmax_weight}
w_f =
\frac{
\exp\left(\frac{(V(f)-\max_{g\in S_{\mathrm{with}}}V(g))}{\tau_V}\right)
}{
\sum_{g\in S_{\mathrm{with}}}
\exp\left(
\frac{(V(g)-\max_{h\in S_{\mathrm{with}}}V(h))}{\tau_V}\right)
}.
\end{equation}
We use \(\tau_V=0.5\). For \(|S_{\mathrm{with}}|<3\), the with-side
aggregate falls back to the arithmetic mean:
\[
    \bar{V}_{\mathrm{with}}
    =
    \mu(S_{\mathrm{with}})
    =
    \frac{1}{|S_{\mathrm{with}}|}
    \sum_{f\in S_{\mathrm{with}}} V(f).
\]
Otherwise, \(\bar{V}_{\mathrm{with}}=\mu_{\mathrm{w}}(S_{\mathrm{with}})\).

\paragraph{Without-side aggregate.}
The without side uses an arithmetic mean, not a softmax-weighted mean:
\[
    \mu(S_{\mathrm{without}})
    =
    \frac{1}{|S_{\mathrm{without}}|}
    \sum_{f\in S_{\mathrm{without}}} V(f),
\]
with \(\mu(S_{\mathrm{without}})=0\) when
\(S_{\mathrm{without}}=\emptyset\). To avoid making the gain undefined or
uninformative when few comparable failures are observed, MSCE shrinks this
mean toward a conservative baseline \(b=0.5\) with pseudocount \(N_0=5\):
\begin{equation}
\label{eq:l2_blend_without}
\bar{V}_{\mathrm{blend}}(S_{\mathrm{without}})
=
\frac{
|S_{\mathrm{without}}|\,\mu(S_{\mathrm{without}})
+
N_0 b
}{
|S_{\mathrm{without}}| + N_0
}.
\end{equation}
Thus, if no without evidence is available,
\(\bar{V}_{\mathrm{blend}}=b\).

\paragraph{Gain.}
The final policy gain is
\[
    G(f^{(2)})
    =
    \bar{V}_{\mathrm{with}}
    -
    \bar{V}_{\mathrm{blend}}(S_{\mathrm{without}}).
\]
This gain is a heuristic utility estimate, not a causal effect estimate.
Softmax weights are used only in gain recomputation; they do not duplicate
or reweight examples inside the L2 induction prompt.

\subsection{L2 Association, Induction, and Revision}
\label{app:l2_details}

For each valued L1 trace, MSCE first attempts association with an existing L2 policy. 
The matching score combines semantic similarity with structured trigger overlap. 
Incompatible domain tags or conflicting error signatures prevent association even when embedding similarity is high. 
Associated traces update the policy's support set and gain estimate.

If no policy matches, the trace is placed into a candidate pool. 
A candidate pool becomes eligible for induction when it contains evidence from at least \(n_{\min}\) distinct episodes. 
This independence requirement prevents a single long trajectory from producing an over-specific policy. 
When new evidence contradicts an active policy, MSCE does not directly overwrite its procedure. 
Instead, the policy gain is recomputed; sustained negative gain retires the policy, while substantial procedural drift triggers re-induction and possible skill rebuilding.

\subsection{L3 Abstraction and Confidence}
\label{app:l3_details}
 
Given active L2 policies, we bucket by a deterministic domain key, admit a cohort via centroid cosine similarity (with a tag-aligned fallback when embeddings diverge), and call the abstraction prompt \(\Pi_{env}\) (Appendix~\ref{app:prompts}) on the cohort plus one capped evidence trace per policy. 
The prompt is constrained to declarative environment knowledge and forbidden from prescribing actions. 
Validated drafts merge into the nearest existing environmental cognition by embedding similarity or instantiate a new environmental cognition. 
Algorithm~\ref{alg:l3} summarizes the procedure.

\begin{algorithm}[t]
\small
\caption{Environmental cognition abstraction}
\label{alg:l3}
\begin{algorithmic}[1]
\REQUIRE active policies $\mathcal{P}$; similarity thresholds $\theta_{\mathrm{sim}}, \theta_{\mathrm{merge}}$; minimum cohort size $m$
\STATE Partition $\mathcal{P}$ into domain buckets $B$
\FOR{each $b \in B$ with $|b| \ge m$}
  \STATE $C \leftarrow \textsc{Admit}(b,\theta_{\mathrm{sim}})$ \COMMENT{strict centroid or tag fallback}
  \STATE $\mathcal{T} \leftarrow$ one recent trace per policy in $C$
  \STATE $d \leftarrow \mathrm{LLM}(\Pi_{\mathrm{env}}, C, \mathcal{T})$
  \IF{$d$ invalid or $\mathrm{conf}(d)=0$} 
  \STATE \textbf{continue} \ENDIF
  \STATE $w^\star \leftarrow \arg\max_{w}\cos(\vec C,\vec w)$
  \IF{$\cos(\vec C,\vec{w^\star}) \ge \theta_{\mathrm{merge}}$}
    \STATE merge $d$ into $w^\star$
  \ELSE
    \STATE insert $d$ as a new environmental cognition
  \ENDIF
\ENDFOR
\end{algorithmic}
\end{algorithm}

The confidence \(c\) of an L3 cognition is increased when newly induced policies support its existing facts and decreased when later evidence contradicts them. 
Low-confidence cognitions are not deleted immediately, since they may still serve as negative or historical evidence; instead, they are excluded from default retrieval until additional support appears.

\subsection{Online Update Operator \texorpdfstring{$\mathcal{U}$}{U}}
\label{app:update_operator}

Algorithm~\ref{alg:update} formalizes the event-driven chain. 
Stages are idempotent per episode identifier; LLM stages are skipped when no provider is configured, while L1 persistence still proceeds.

\begin{algorithm}[t]
\small
\caption{Online update operator $\mathcal{U}$}
\label{alg:update}
\begin{algorithmic}[1]
\STATE \textbf{Upon} episode finalization:
\STATE \quad Extract L1 steps
\STATE \quad Truncate the text
\STATE \quad Extract or synthesize self-reflection \(\rho\)
\STATE \quad Score $\alpha$ via $\Pi_{\mathrm{reflexion\_score}}$
\STATE \quad Embed summaries; 
\STATE \quad Write traces with $V=\textsc{pending}$.
\STATE \textbf{Upon} receipt of $R_i$ (feedback or timeout):
\STATE \quad Build an episode summary
\STATE \quad $R_i \leftarrow$ environment reward or
\(\mathrm{LLM}(\Pi_{\mathrm{reward}}, \cdot)\)
\STATE \quad Backfill $V$
\STATE \quad Decay retrieval priorities
\STATE \textbf{Upon} reward update:
\quad \FOR{traces with $V \ge v_{\min}$}
    \STATE \quad Associate or pool
    \STATE \quad Induce if quorum met
    \STATE \quad Recompute $(G,\text{support},\text{status})$
\quad \ENDFOR
\STATE \textbf{Upon} active L2 change:
\STATE \quad Abstract L3 clusters
\STATE \quad Evaluate skill eligibility
\STATE \quad Crystallize $\rightarrow$ verify $\rightarrow$ probationary deploy
\STATE \textbf{Upon} failure burst or user correction:
\STATE \quad Synthesize contrastive guidance 
\STATE \quad Append to $\mathcal{D}$
\STATE \quad Stash repair packet for the next turn
\STATE \textbf{On failure:} 
\STATE \quad Malformed JSON $\Rightarrow$ skip stage 
\STATE \quad Empty evidence $\Rightarrow$ no new skill
\end{algorithmic}
\end{algorithm}

\subsection{Skill Lifecycle}
\label{app:skill_lifecycle}

Skills pass through three states: probationary, active, and archived. 
A newly crystallized skill is probationary. 
It becomes active when its reliability \(\eta\) exceeds a threshold after sufficient invocations. 
It is archived when repeated failures, explicit user rejection, or negative source-policy gain indicate that it should no longer be retrieved by default.

Reliability is estimated by a smoothed success rate:
\[
    \eta =
    \frac{n_{\mathrm{pass}}+1}{n_{\mathrm{trial}}+2}.
\]
In addition, explicit positive or negative user feedback can adjust \(\eta\), and boundary violations shrink the applicability boundary \(\mathcal{B}\). 
When the source L2 policy is substantially revised, the corresponding skill is rebuilt from a fresh evidence set rather than patched in place. The detailed lifecycle operations for crystallized skills are provided in Table~\ref{tab:skill_lifecycle}.

\begin{table}[t]
\centering
\small
\caption{Lifecycle operations for crystallized skills.}
\label{tab:skill_lifecycle}
% \resizebox{\columnwidth}{!}
{
\begin{tabular}{lll}
\toprule
\textbf{Event} & \textbf{Operation} & \textbf{Effect} \\
\midrule
successful invocation & reinforce & increase reliability \\
execution failure & repair & revise procedure or verification \\
user rejection & shrink & narrow applicability boundary \\
new counter-evidence & revise & add guidance or anti-patterns \\
source policy rewritten & rebuild & re-crystallize from evidence \\
long inactivity or low reliability & archive & remove from default retrieval \\
\bottomrule
\end{tabular}
}
\end{table}

\subsection{Hierarchical Retrieval}
\label{app:hierarchical_retrieval}

At inference time, MSCE retrieves memory in three tiers.

\paragraph{Skill retrieval.}
The router first matches the current context against active skill triggers. 
Matched skills are ranked by trigger relevance, reliability \(\eta\), and source-policy gain. 
The selected skill contributes its procedure, preconditions, verification rule, boundary, evidence anchors, and decision guidance to the agent context.

\paragraph{Trace and episode retrieval.}
If no skill applies, or if a skill fails, MSCE retrieves L1 evidence. 
Trace retrieval combines exact cues such as error signatures with semantic similarity over compact state summaries. 
High-value traces are prioritized, while low-value traces are preserved for counter-evidence and decision repair. 
% When several retrieved traces belong to the same task, they may be presented as an task-level reference plan.

\paragraph{Environmental cognition retrieval.}
When the agent expresses structural uncertainty, such as not knowing where files, tests, configurations, or constraints are located, MSCE retrieves relevant L3 environmental cognition. 
The retrieved L3 cognition provides environmental priors but does not directly prescribe actions.

\subsection{Decision Guidance and Repair}
\label{app:decision_repair}

MSCE converts repeated contrastive evidence into decision guidance \(\mathcal{D}\). 
When two action patterns occur in similar contexts but lead to substantially different values, and the lower-valued pattern is supported by failures or user corrections, MSCE creates a guidance item:
\[
    d = (c, a^{+}, a^{-}, e, \xi),
\]
where \(c\) is the context, \(a^{+}\) is the preferred action, \(a^{-}\) is the action to avoid, \(e\) contains evidence links, and \(\xi\) is guidance reliability. 
Guidance is injected only when its context matches the current skill or policy trigger. 
Later feedback can increase, decrease, or retire \(\xi\).

\subsection{Storage and Privacy Controls}
\label{app:privacy}

Because L1 traces may contain user text, tool outputs, file paths, or environment-specific information, MSCE applies bounded persistence. 
Long fields are truncated, repeated adjacent steps are deduplicated, and sensitive strings are removed through rule-based redaction before storage. 
Higher-level memories store evidence identifiers rather than raw unbounded observations. 
These controls reduce memory bloat and limit propagation of sensitive content, although they do not eliminate all privacy risks.

\begin{table*}[t]
\centering
\small
\caption{L1 traces from two distinct tasks demonstrating cross-task L2 policy induction. Task-level feedback (\(R_A = 0.8\), \(R_B = 0.9\)) is backpropagated using reflection weights (\(\alpha\)) to compute step values (\(V\)).}
\label{tab:example_traces}
\resizebox{\textwidth}{!}{
\begin{tabular}{clllllcc}
\toprule
\textbf{Task} & \textbf{Trace} & \textbf{Environment} & \textbf{Action} & \textbf{Observation} & \textbf{Reflection (\(\rho\))} & \textbf{\(\alpha\)} & \textbf{\(V\)} \\
\midrule
\multirow{2}{*}{A} & \(f_{1,1}^{(1)}\) & Alpine Docker & \texttt{pip install lxml} & \texttt{Error: xmlsec1 not found} & ``Alpine lacks C library xmlsec1'' & 0.7 & 0.776 \\
& \(f_{1,2}^{(1)}\) & Alpine Docker & \texttt{apk add... \&\& pip...} & Success & ``Needs system lib before compile'' & 0.5 & 0.80 \\
\midrule
\multirow{2}{*}{B} & \(f_{2,1}^{(1)}\) & Debian Docker & \texttt{pip install psycopg2} & \texttt{Error: pg\_config not found} & ``Missing PostgreSQL dev library'' & 0.6 & 0.864 \\
& \(f_{2,2}^{(1)}\) & Debian Docker & \texttt{apt-get... \&\& pip...} & Success & ``Strategy effective'' & 0.4 & 0.90 \\
\bottomrule
\end{tabular}
}
\end{table*}

\section{Example: Environment Dependencies}
\label{sec:example}

To illustrate the MSCE workflow more clearly, consider an agent resolving environment dependencies across two distinct tasks shown in Table~\ref{tab:example_traces}.

In \textbf{Task A} (``Install \texttt{lxml} in Alpine''), the agent's initial \texttt{pip install} fails. The agent reflects on the missing C-library, installs it via the package manager, and succeeds. The user's positive feedback \(R_A = 0.8\) is backpropagated using the reflection weights \(\alpha\), assigning high trace values \(V\) to these exploratory steps. For instance, assuming a discount factor \(\gamma=0.9\), the value of the key discovery step \(f_{1,1}^{(1)}\) is computed via reflection-weighted backfilling as \(V = 0.7 \times 0.8 + (1 - 0.7) \times 0.9 \times 0.8 = 0.776 \). Since no existing L2 policy matches, these L1 traces enter the candidate pool. 

Later, in \textbf{Task B} (``Deploy Django App''), a similar failure occurs with \texttt{psycopg2} on a Debian-based container. The user confirms the resolution, yielding another set of high-value traces.

Upon the arrival of Task B's traces, the online update operator \(\mathcal{U}\) detects a cross-task overlap (container environment, \texttt{pip} compilation failure, missing system libraries) between \(f_{1,1}^{(1)}\) and \(f_{2,1}^{(1)}\). This triggers the induction of an \textbf{L2 policy}:
\begin{itemize}
    \item \textbf{Trigger:} \texttt{pip install} fails due to missing system libraries in a container.
    \item \textbf{Procedure:} Parse the missing component \(\rightarrow\) identify the OS package manager (e.g., \texttt{apk} or \texttt{apt-get}) \(\rightarrow\) install the corresponding \texttt{-dev} library \(\rightarrow\) retry \texttt{pip install}.
    \item \textbf{Boundary:} Containers only; excludes native systems with pre-installed libraries. 
\end{itemize}

As this L2 policy demonstrates stable positive gain, MSCE abstracts an \textbf{L3 environmental cognition} (e.g., the three-layer dependency structure of Python C-extensions) and crystallizes the policy into a \textbf{skill}, avoiding redundant trial-and-error in future tasks.

\section{Experimental Details}

\subsection{Experimental Settings}\label{app:settings}

\paragraph{Evaluation Protocol}

On EvoAgentBench, for task~$i$, the per-task success rate is
$\mathrm{Pass@1}_i=\frac{1}{n_i}\sum_{j=1}^{n_i}\mathbb{I}[r_{i,j}>\tau]$.
The benchmark-level Pass@1 is the unweighted mean over tasks,
$\mathrm{Pass@1}=\frac{1}{N}\sum_{i=1}^{N}\mathrm{Pass@1}_i$,
with standard error computed across tasks. Regarding how reward $r$ is assigned, we follow the EvoAgentBench settings. Information Retrieval uses LLM-as-a-judge with exact-match fallback ($r\in\{0,1\}$).
Reasoning uses normalized exact matching or optional LLM-based mathematical equivalence ($r\in\{0,1\}$).
Software Engineering follows the official SWE-bench \texttt{resolved} criterion ($r\in\{0,1\}$).
Code Implementation requires passing all hidden test cases ($r\in\{0,1\}$).
Knowledge Work uses LLM rubric scoring with a ClawWork-compatible cliff: scores below $0.6$ are mapped to failure, and success requires $r>0.6$.

\paragraph{Hyperparameters}

Table~\ref{tab:hparams} lists the principal hyperparameters used in our implementation. 
These values are not intrinsic to MSCE and can be tuned for different environments.

\begin{table}[h]
\centering
\small
\caption{Principal MSCE hyperparameters.}
\label{tab:hparams}
% \resizebox{\columnwidth}{!}
{
\begin{tabular}{lll}
\toprule
\textbf{Symbol} & \textbf{Meaning} & \textbf{Value} \\
\midrule
\multicolumn{3}{l}{\textit{Trace valuation and reward}} \\
\(\gamma\) & value backfilling discount & 0.9 \\
\(w_g\) & reward weights for goal & 0.45 \\
\(w_p\) & reward weights for process & 0.30 \\
\(w_u\) & reward weights for satisfaction & 0.25 \\
\midrule
\multicolumn{3}{l}{\textit{L2 policy induction and gain}} \\
\(n_{\min}\) & distinct episodes for L2 induction & 2 \\
\(v_{\min}\) & minimum value for L2 association & 0.1 \\
\(\tau_V\) & temperature for value weighting in policy gain & 0.5 \\
\(N_0\) & pseudo-count in policy gain & 5 \\
\(b\) & shrinkage anchor in policy gain & 0.5 \\
\(\theta_G\) & gain threshold for skill promotion & 0 \\
\midrule
\multicolumn{3}{l}{\textit{L3 environmental cognition abstraction}} \\
\(m\) & minimum policy cohort size for L3 abstraction & 2 \\
\(\theta_{\mathrm{sim}}\) & cohort admission similarity threshold & 0.62 \\
\(\theta_{\mathrm{edge}}\) & related-topic Jaccard edge threshold  & 0.34 \\
\midrule
\multicolumn{3}{l}{\textit{Skill crystallization and lifecycle}} \\
\(K\) & maximum evidence traces per skill & 6 \\
\(k\) & retrieved skills per task & 3 \\
\(n_{\mathrm{prob}}\) & minimum probationary invocations before activation & 1 \\
\(\theta_{\eta}^{\mathrm{active}}\) & reliability threshold for active retrieval & 0.6 \\
\(\theta_{\eta}^{\mathrm{archive}}\) & reliability threshold for archiving & 0.2 \\
\(\delta\) & explicit-feedback update step for \(\eta\) & 0.1 \\
\midrule
\multicolumn{3}{l}{\textit{Evaluation}} \\
\(\tau_{\mathrm{eval}}\) & success threshold for Pass@1 unless KW & 0; 0.6 for KW \\
\bottomrule
\end{tabular}}
\end{table}

\subsection{Effect of Simulated Human Feedback}
\label{app:simulated_human_feedback}

MSCE is designed to use not only numerical environment rewards but also textual user feedback. 
To examine whether textual episode-level feedback provides useful learning signals, we conduct an additional controlled study on EvoAgentBench with LLM-simulated human feedback. 
This experiment also provides an indirect validation of the reward quantification mechanism in Appendix~\ref{app:reward_shaping}, since the simulated feedback is converted into scalar terminal feedback by the same quantification module used in MSCE.

\paragraph{Setup.}
We compare two configurations. 
\textbf{C0} denotes the default MSCE setting without injected simulated human feedback. 
\textbf{C1} augments episode-level feedback with simulated human feedback generated by an LLM mentor (Opus 4.7 Max). 
For failed or blocked training trajectories, the mentor observes the task prompt, agent trajectory, failure or blocker signal, verifier result, and existing memory/reflection content, and produces a concise root-cause-aligned feedback message. 
The feedback is written in the same style and format as existing memory or reflection content, so that it does not introduce a new memory schema or additional test-time hints. 
This feedback is injected only during the online evolving phase and is then quantified into terminal feedback \(R_i\) using the reward quantification mechanism in Appendix~\ref{app:reward_shaping}. 
At evaluation time, no simulated feedback is generated; the agent uses only the memory and skills accumulated during training.

\paragraph{Feedback generation prompt.}
The LLM mentor is instructed to behave like an experienced human coach rather than an oracle. 
It must not provide final answers, complete code, or task-specific shortcuts. 
Instead, it identifies what the agent was trying to do, where the reasoning or action deviated from the correct path, why the deviation caused failure or blockage, what general principle should be remembered, and which concrete evidence or check should be prioritized in similar future tasks. 
This design encourages feedback that is reusable as training memory rather than direct supervision for a particular evaluation instance.

\paragraph{Results.}
Table~\ref{tab:simulated_human_feedback} compares \textbf{C0} and \textbf{C1}.
Simulated feedback improves Pass@1 on four of the five domains, with the largest gains on KW and SE, where \textbf{C1} improves Pass@1 by \(+13.79\) and \(+7.69\) points, respectively.
Code and IR also improve by \(+2.56\) and \(+1.54\) points, while Math remains unchanged, suggesting that its bottleneck may lie more in exact symbolic correctness than in procedural improvement driven by richer feedback.

The cost trend is mixed: \textbf{C1} reduces cost on Code, IR and SE but increases cost on KW, and Math.
This suggests that richer feedback can reduce redundant exploration in some tool-use tasks, while encouraging more verification or elaboration in others.
Overall, these results suggest that episode-level textual feedback can provide useful additional supervision for MSCE's memory--skill evolution, and that the proposed reward quantification mechanism can convert such feedback into effective update signals.

\begin{table}[t]
\centering
\small
\setlength{\tabcolsep}{3.2pt}
\renewcommand{\arraystretch}{1.08}
\caption{
Effect of LLM-simulated human feedback on EvoAgentBench.
C0 denotes MSCE without simulated feedback, and C1 denotes MSCE with simulated feedback injected during training.
Cost is measured in chars for Math and turns for other domains, so cost comparisons should be made within each domain.
}
\label{tab:simulated_human_feedback}
% \resizebox{\columnwidth}{!}
{
\begin{tabular}{lccc|ccc}
\toprule
\multirow{2}{*}{\textbf{Domain}}
& \multicolumn{3}{c|}{\textbf{Pass@1} $\uparrow$}
& \multicolumn{3}{c}{\textbf{Cost} $\downarrow$} \\
\cmidrule(lr){2-4}
\cmidrule(lr){5-7}
& \textbf{C0} & \textbf{C1} & \(\Delta\)
& \textbf{C0} & \textbf{C1} & \(\Delta\) \\
\midrule
Code 
& 61.54 & 64.10 & +2.56 
& 1.97 & 1.74 & -0.23 \\
SE 
& 53.85 & 61.54 & +7.69 
& 40.80 & 37.71 & -3.09 \\
KW
& 53.45 & 67.24 & +13.79
& 14.97 & 16.90 & +1.93 \\
Math
& 47.00 & 47.00 & +0.00 
& 1140.9 & 1150.0 & +9.1 \\
IR 
& 26.15 & 27.69 & +1.54 
& 8.74 & 8.40 &  -0.34 \\
\bottomrule
\end{tabular}}
\end{table}

\section{Prompted Operators}
\label{app:prompts}

MSCE uses LLMs for five evidence-constrained operators: reflection scoring, reward quantification, L2 policy induction, and L3 environmental cognition abstraction, and skill drafting. 
All prompted operators are required to return structured fields and evidence identifiers. 
Outputs that lack required fields, refer to unsupported evidence, or introduce unsupported tools are discarded rather than inserted into memory.

\subsection{Reflection Scoring}\label{app:reflection_score}

Given a step context \((s_{i,t}, a_{i,t}, o_{i,t}, \rho_{i,t})\), the reflection scorer estimates \(\alpha_{i,t}\in[0,1]\). 
The rubric considers four aspects: faithfulness to the observed step, causal insight, transferability to future tasks, and concreteness. 
Empty or tautological reflections receive \(\alpha_{i,t}=0\). If reflection
scoring is disabled, non-empty reflections receive a neutral value
\(\alpha_{i,t}=0.5\); malformed judge outputs fall back to the existing
reflection metadata rather than updating the trace.
In batched mode, all steps in a short task are scored jointly so that the judge can compare their relative contribution. The specific prompt (\(\Pi_{\mathrm{reflexion\_score}}\)) is provided as follows:
\begin{lstlisting}
You are a strict reviewer of agent self-reflections.

You see the FULL context of one agent step:
- STATE        — what the agent saw before acting (user prompt, prior observation)
- THINKING     — the LLM's own native chain-of-thought for this step, if any (Claude extended-thinking, pi-ai ThinkingContent). Empty when the model didn't emit thinking this turn.
- ACTION       — what the agent produced (assistant text output)
- TOOL_CALLS   — tools the agent invoked this step, with inputs and outputs (or errors). Tool usage + outcomes are part of the action chain and carry their own signal about what the agent did.
- OUTCOME      — the final observable result of the step (last tool outcome or "(assistant-only step)" for pure text turns)
- REFLECTION   — the text being graded: the agent's first-person explanation of WHY it acted this way and WHAT it learned.

Score the REFLECTION on four axes, combined into ONE number α ∈ [0, 1]:

  1. faithfulness  — does the reflection match what ACTUALLY happened across THINKING + ACTION + TOOL_CALLS + OUTCOME?
  2. causal insight — does it identify why the action / tool choice worked or failed? Bonus when it connects the model's visible THINKING to the resulting action.
  3. transferability — does it surface a lesson useful on a similar future task?
  4. concreteness  — are the details specific (real command names, real error messages, real decisions) rather than generic platitudes like "I should do better"?

Rules:
- THINKING and TOOL_CALLS are first-class evidence for grading α — a reflection that ignores a visible thinking chain or misreports a tool call should score LOW on faithfulness.
- TOOL_CALLS that errored are strong signal: the reflection should name the error and what it implied. Missing that is a faithfulness penalty.
- An empty / purely-tautological reflection → α = 0, usable = false.
- α ≥ 0.4 AND reflection non-tautological → usable = true; else false.

Return JSON:
{
  "alpha": 0.0-1.0,
  "usable": true | false,
  "reason": "one-sentence justification"
}
\end{lstlisting}

\subsection{Reward quantification.}
When no numerical environment reward is available, textual user feedback is converted into \(R_i\). 
The evaluator scores goal achievement, process quality, and user satisfaction. 
The input includes the original query \(q_i\), the final output \(\hat{y}_i\), a compact episode summary, and user feedback \(h_i\). The specific reward quantification prompt (\(\Pi_{\mathrm{reward}}\)) is as follows:

\begin{lstlisting}
You are a strict grader of AI-agent task execution.

You receive:
- TASK_SUMMARY  — the FULL conversation arc for this task: * USER_ASKS_AND_AGENT_REPLIES lists every user turn paired with the agent's corresponding reply, in chronological order. One "task" frequently spans multiple user turns as the user refines / follows up / pivots topics within the same session. * MOST_RECENT_USER_ASK and MOST_RECENT_AGENT_REPLY call out the final exchange explicitly — that is usually the truest signal of whether the agent is actually tracking where the user is now.
- FEEDBACK       — the user's own messages AFTER the task attempt finished. May be short ("ok thanks"), explicit ("try again with X"), or structured ("resolved, but too slow"). Frequently empty.

Grade the agent on THREE INDEPENDENT AXES, each in [-1, 1]:

1. "goal_achievement" — did the agent address what the user ACTUALLY asked?
   +1.0  every user ask across the exchange was addressed correctly.
   +0.3  the last ask was addressed well; earlier asks had minor gaps.
   0.0   unclear if the user's ask was met.
   -0.3  missed a significant portion of what was asked.
   -1.0  fundamentally wrong answer / caused damage.

   CRITICAL RULE — do NOT anchor on the first user turn. A user who starts with "上海天气" and later pivots to "再查北京天气" is a user whose goal has EVOLVED; if the agent answered Beijing on the final turn when asked about Beijing, that is goal-achievement = POSITIVE, not negative. Judge each user ask on its own merits, weighted toward the most recent exchange (which is where the user actually is now).

2. "process_quality"
   +1.0  clean, minimal, correct reasoning across all turns.
   0.0   reasonable but not great.
   -1.0  lots of thrashing, wrong tools, noisy output.

3. "user_satisfaction"  (from FEEDBACK text tone + trailing user asks)
   +1.0  thanks / happy / "做的很好" / accepts and closes out.
   +0.3  moves on neutrally to next ask or new topic.
   0.0   no emotional signal either way.
   -0.3  asks for correction ("no, do X instead" / "重做").
   -1.0  hard-stops, expresses frustration.

Rules:
- If FEEDBACK is empty, infer satisfaction CONSERVATIVELY from the last exchange's tone. A follow-up question is usually ≈ 0 (neutral continuation), NOT negative. Never invent anger.
- Base scores ONLY on what TASK_SUMMARY actually describes — do not assume facts not shown.
- You are grading the HOST AGENT described in HOST_AGENT_CONTEXT, not yourself. Do NOT use your own model identity, provider, policies, or capabilities to decide whether the host agent answered identity/model questions correctly. If hostModel/hostProvider are provided, treat them as the authoritative runtime context unless the conversation itself contains a correction.
- Produce one short justification.

Return JSON, EXACTLY this shape (no extra keys, no commentary):
{
  "goal_achievement":  number in [-1, 1],
  "process_quality":   number in [-1, 1],
  "user_satisfaction": number in [-1, 1],
  "label": "success" | "partial" | "failure" | "unknown",
  "reason": "one-sentence justification"
}
\end{lstlisting}

\subsection{L2 induction.}
The L2 induction prompt receives evidence traces selected from distinct episodes. 
Each evidence item contains a compact state, action, outcome, value \(V\), reflection weight \(\alpha\), and evidence id. 
The model is asked to produce a trigger, procedure, verification rule, applicability boundary, and supporting evidence ids. 
The prompt (\(\Pi_{\mathrm{policy}}\))is explicitly restricted to procedural knowledge: declarative environment facts are deferred to L3 abstraction as follows:

\begin{lstlisting}
You induce reusable **procedural policies** from agent experience.

A policy is a "how-to": "when you see condition X in the agent's state, do action Y, verify with Z, watch out for caveat W." It is **NOT** a description of the environment.

Input TRACES: a list of { state_summary, action, outcome, utility } records that all share a similar state signature.

Produce ONE policy describing the action pattern. The policy must:
- Name a TRIGGER recognizable from the agent's STATE — a condition the agent can detect at the moment of decision (an error code, a missing file, a request shape). NOT a fact about the environment in general.
- Prescribe an ACTION template — a parameterized step or short step sequence. Templates over single exact commands. NOT a single example.
- Note at least one CAVEAT or failure mode observed in the traces — a step-level pitfall, NOT a generic environment taboo.
- Generalize across the input traces, not restate one of them.

──────────────────── Boundaries — what NOT to write ────────────────────

This output is a **procedural policy**, not environmental cognition. Environmental cognition lives in a separate layer (L3) generated by a different prompt. Cross-contamination on either side dilutes both.

Do NOT write any of these — they belong to L3 (environmental cognition), not here:
  - Topology facts: "Alpine containers ship musl libc" "Python deps form a 3-layer stack" "src/components/ holds React components"
  - Environment behavioural rules (in pure declarative form): "binary wheels are incompatible with musl" "the service reads config only at startup"
  - Environment taboos detached from a specific action choice: "this directory is read-only" "production tables shouldn't be DROPped lightly"

If a trace tells you the environment looks a certain way, FOLD that fact INTO the trigger or caveat as a state-level CONDITION the agent can check, not as a standalone description. Example:

  Wrong (drifts into env-fact):
    trigger:  "Alpine ships musl libc"
    caveats:  ["Python deps have a 3-layer stack"]

  Right (states it as actionable conditions):
    trigger:  "container is Alpine AND pip install fails with '<lib> not found' or 'header not found'"
    caveats:  ["if first apk add still fails, also check musl-vs-glibc wheel compatibility before retrying"]

──────────────────── Same fact, two framings ─────────────────────

If the underlying truth is "Alpine containers don't ship system dev libs by default":

  Express here (procedural): "When pip install fails inside an Alpine container with a missing system library, run apk add <pkg>-dev then retry pip."

  Do NOT express here (declarative — that's L3's job): "Alpine container images ship only the pure-Python tier of the Python dependency stack."

──────────────────── Output ─────────────────────

Return JSON:
{
  "title": "short imperative title",
  "trigger": "state-level condition the agent can detect",
  "action": "templated step or step sequence",
  "rationale": "why this action works ON THESE TRACES (not why the
                environment behaves this way)",
  "caveats": ["step-level pitfall string", ...],
  "confidence": number in [0, 1],
  "support_trace_ids": ["tr_...", ...]
}
\end{lstlisting}

\subsection{L3 abstraction.}
The L3 abstraction prompt receives a cohort of active L2 policies and representative evidence. 
It extracts declarative environment knowledge in three fields: entities or structures \(\mathcal{E}\), action--response regularities \(\mathcal{I}\), and constraints \(\mathcal{C}\). 
The prompt (\(\Pi_{\mathrm{env}}\)) forbids imperative procedural instructions, so that L3 remains an environmental cognition rather than a skill as follows:

\begin{lstlisting}
You abstract environmental cognition from cross-task policy evidence.

An environmental cognition model is **declarative** knowledge about how the environment IS:
its topology, its causal/behavioural regularities, its taboos. It is **NOT** a recipe for what to do — that lives in the L2 procedural layer, generated by a separate prompt. Cross-contamination on either side dilutes both.

Input POLICIES: a list of L2 policies (with trigger / procedure /verification / boundary / support / gain), plus a short sample of the L1 traces that minted each. Every policy shares a compatible domain (matched by primary tag / tool).

Produce ONE environmental cognition model describing the **environment** those policies
operate in. It must answer:

  - Environment topology (ℰ) — what lives where, what is the shape of this environment? Pure facts of existence and structure.
    GOOD: "Alpine containers ship musl libc, no glibc"
          "Node project repos group source under src/"
          "macOS bundles BSD sed; Linux distros bundle GNU sed"
    BAD (drifts into procedure):
          "use apk add to install system libs"
          "prefer Python scripts over sed on macOS"

  - Inference rules (ℐ) — how does the environment causally respond to common stimuli? Phrase as cause→effect, NOT as guidance.
    GOOD: "loading a glibc-linked binary wheel inside Alpine raises a dynamic-link error" "editing config.yaml does not propagate until the process restarts (no in-process watcher)"
    BAD (drifts into procedure): "if pip install fails, install dev libs and retry" ← that's an action plan, belongs to L2 "always restart the service after editing config" ← that's a recommendation, belongs to L2

  - Constraints (C) — what facts of the environment make some actions unsafe or invalid? State the FACT, not the avoidance behavior.
    GOOD: "node_modules/ is rewritten by npm install; manual edits are lost on the next sync" "production database tables hold customer data; destructive DDL is irreversible"
    BAD (drifts into procedure): "don't edit node_modules/ directly" ← that's a behavioural rule, belongs to L2 / decision repair "don't run DROP TABLE in production" ← same — phrase the underlying environment fact instead

Do NOT, under any section:
  - Use imperative or recommendation verbs (do / don't / should / use / prefer / avoid / try / install / run). The environmental cognition never tells the agent what to do.
  - Restate a single trace — the cognition model must generalise across policies.
  - Include advice tied to a single user or session.

──────────────────── Same fact, two framings ─────────────────────

If the underlying truth is "Alpine containers don't ship system dev
libs by default":

  Express here (declarative):
    inference: "Python C-extension packages fail to compile in Alpine
                containers when the matching system header / library
                package is not pre-installed in the image."

  Do NOT express here (procedural — that's L2's job):
    "When pip fails in Alpine, apk add <pkg>-dev and retry pip."

──────────────────── Output ─────────────────────

Return JSON:
{
  "title": "short noun phrase, e.g. 'Alpine python dependency model'",
  "domain_tags": ["tag1", "tag2"],   // 1-4 short, lowercase, no spaces
  "environment": [
    { "label": "...", "description": "...", "evidenceIds": ["po_...", "tr_..."] }
  ],
  "inference":   [ { "label": "...", "description": "...", "evidenceIds": [] } ],
  "constraints": [ { "label": "...", "description": "...", "evidenceIds": [] } ],
  "body": "rendered markdown summary of the three sections",
  "confidence": number in [0, 1],
  "supersedes_cognition_ids": []          // optional: prior cognition this refines
}
\end{lstlisting}

\subsection{Skill drafting.}
The skill crystallization prompt receives the source L2 policy, ranked
supporting evidence, optional counter-examples, a whitelist of observed
tools, decision-guidance seeds, and the existing skill naming space.
It returns a callable procedure with preconditions, ordered steps, examples, and decision guidance. 
The generated skill is accepted only if its tools and claims are supported by the evidence set.
The specific prompt (\(\Pi_{\mathrm{skill}}\)) is as follows:

\begin{lstlisting}
You crystallize a skill an agent should be able to call.

Input:
- POLICY: the L2 policy being promoted (trigger / action / rationale / caveats).
- EVIDENCE: 3..10 successful traces that support the policy.
- EVIDENCE_TOOLS: the exhaustive list of tool/command names that actually appeared in the evidence traces' tool calls. This is the ground-truth
  whitelist — your `tools` output MUST be a subset of this list.
- COUNTER_EXAMPLES (optional): traces with V < 0 from the same context —
  failures the policy is meant to prevent.
- REPAIR_HINTS (optional): a JSON block { preference: [...], antiPattern: [...] } attached to the policy by the decision-repair pipeline. These are concrete "prefer / avoid" lines synthesised from earlier failures + user feedback; treat them as authoritative seeds for `decision_guidance` below.
- NAMING_SPACE: a list of existing skill names to avoid colliding with.

Return JSON:
{
  "name": "snake_case_identifier, ≤ 32 chars, unique vs NAMING_SPACE",
  "display_title": "human title in user's language",
  "summary": "2-3 sentence description of what the skill does and when to use it",
  "parameters": [
    { "name": "...", "type": "string|number|boolean|enum", "required": true|false,
      "description": "...", "enum": ["..."] }
  ],
  "preconditions": ["bullet", ...],
  "steps": [
    { "title": "short", "body": "markdown-friendly paragraph describing the step" }
  ],
  "examples": [
    { "input": "...", "expected": "..." }
  ],
  "tools": ["tool_or_command_name", ...],
  "decision_guidance": {
    "preference":   ["Prefer: …", ...],   // concrete actions to favour, ≤ 5
    "anti_pattern": ["Avoid: …", ...]     // concrete actions to avoid, ≤ 5
  },
  "tags": ["optional string", ...]
}

Rules:
- `tools` MUST only contain names from EVIDENCE_TOOLS. Never invent tool names that are not in the whitelist. Include every tool the skill's procedure actually invokes — omit tools not referenced in your steps.
- Keep "steps" short (2-6 items).
- `summary` must be self-contained so the agent can decide whether to call this skill without reading the full SKILL.md.
- For `decision_guidance`:
  - If REPAIR_HINTS is non-empty, fold each line in verbatim (or lightly normalised) — they are already grounded in evidence and user feedback.
  - You MAY add 1–2 extra entries derived from contrasting EVIDENCE (high-V) vs COUNTER_EXAMPLES (low-V), if they materially clarify the decision. Don't invent guidance unsupported by the inputs.
  - Each entry should be one short, actionable sentence (≤ 200 chars).
  - Empty arrays are fine when there's nothing to say — never fabricate.
\end{lstlisting}

\end{document}